\pdfoutput=1

\documentclass[11pt]{article}

\usepackage[preprint]{acl}

\usepackage{times}
\usepackage{latexsym}

\usepackage[T1]{fontenc}

\usepackage[utf8]{inputenc}

\usepackage{microtype}

\usepackage{inconsolata}

\usepackage{graphicx}
\usepackage{hyperref}
\usepackage{float}
\usepackage{color}
\usepackage{xcolor}
\usepackage{colortbl}
\usepackage{multirow}
\usepackage{array}
\usepackage{threeparttable}
\usepackage{authblk}
\usepackage[symbol]{footmisc}
\usepackage{multicol}
\usepackage{booktabs}
\usepackage{subcaption}
\usepackage{adjustbox}
\usepackage{algorithm}
\usepackage{algpseudocode}
\usepackage{amsmath}
\usepackage{soul}
\usepackage{bbding}
\usepackage[frozencache,cachedir=minted-cache]{minted}
\usepackage{longtable}
\usepackage[most]{tcolorbox}
\usepackage{enumitem}
\usepackage{titlesec}

\definecolor{Gray}{gray}{0.95}
\definecolor{aigold}{RGB}{244,210, 1} 
\definecolor{aigreen}{RGB}{210,244,211} 

\setlist[itemize]{leftmargin=4mm,itemsep=0mm}
\titlespacing{\paragraph}{%
  0pt}{
  0.2\baselineskip}{
  1em}

\newcommand{\eat}[1]{\ignorespaces}

\newcommand{\Hquad}{\hspace{0.7em}}

\title{Benchmarking Data Science Agents}
\author{%
  Yuge Zhang$^1$ \Hquad Qiyang Jiang$^2$ \Hquad Xingyu Han$^2$ \Hquad Nan Chen$^1$ \Hquad Yuqing Yang$^1$ \Hquad Kan Ren$^2$ \thanks{Correspondence to Kan Ren.} \\
  \textsuperscript{\rm 1}Microsoft Research, \textsuperscript{\rm 2}ShanghaiTech University \\
  \texttt{Yuge.Zhang@microsoft.com, renkan@shanghaitech.edu.cn}
}

\begin{document}

\maketitle

\begin{abstract}

In the era of data-driven decision-making, the complexity of data analysis necessitates advanced expertise and tools of data science, presenting significant challenges even for specialists. 
Large Language Models (LLMs) have emerged as promising aids as data science agents, assisting humans in data analysis and processing. 
Yet their practical efficacy remains constrained by the varied demands of real-world applications and complicated analytical process.
In this paper, we introduce DSEval -- a novel evaluation paradigm, as well as a series of innovative benchmarks tailored for assessing the performance of these agents throughout the entire data science lifecycle.
Incorporating a novel bootstrapped annotation method, we streamline dataset preparation, improve the evaluation coverage, and expand benchmarking comprehensiveness.
Our findings uncover prevalent obstacles and provide critical insights to inform future advancements in the field.\footnote{Source code and data are available at \url{https://github.com/MetaCopilot/dseval}.}

\end{abstract}

\section{Introduction}

Data science has become significant, as it helps individuals and organizations make informed decisions, predict trends, and improve processes by analyzing large volumes of data.
Research on this topic continues to advance the field, driving innovations in machine learning, artificial intelligence, and big data analytics, thus enhancing its impact across various industries.
However, data science requires extensive knowledge about analytical toolkits (e.g., NumPy and Pandas) and professional expertise to conduct analysis and correctly draw insights from data, which is challenging even for specialists.

Recent advancements in Large Language Model (LLM) \cite{brown2020language,touvron2023llama} and LLM-powered agents \cite{shen2023hugginggpt} have shown considerable potential in enhancing human capabilities in data science.
For instance, Code Interpreter\footnote{\url{https://openai.com/blog/chatgpt-plugins}} allows ChatGPT to perform data analysis and visualization by creating a sandboxed Python interpreter within the platform.
Copilots integrated with Microsoft Excel and PowerBI\footnote{\url{https://support.microsoft.com/en-us/copilot-excel}} assist users in exploring and understanding data and finding insights.
Similar initiatives have also emerged in the open-source community, such as Jupyter AI~\cite{jupyterai}, Chapyter~\cite{chapyter}, and CoML~\cite{zhang2023mlcopilot}.

\begin{figure}[t]
\includegraphics[width=\linewidth,page=4]{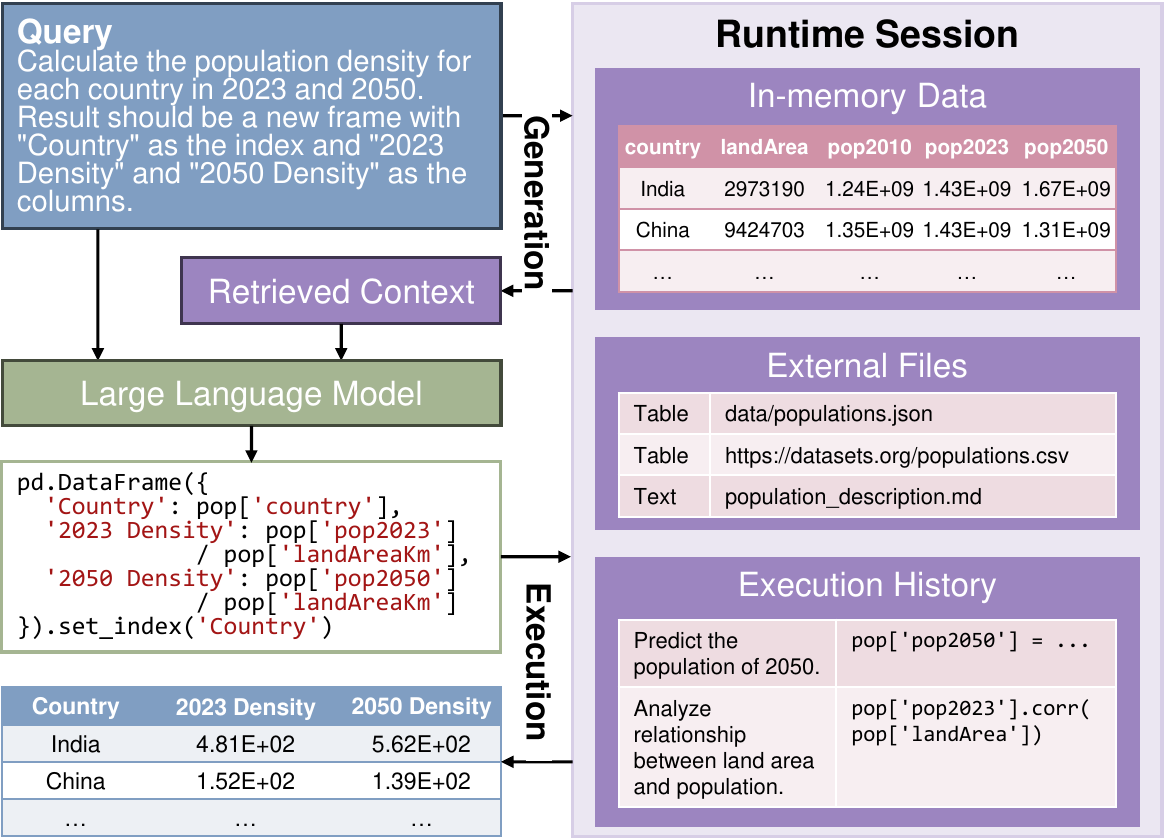}
\caption{Illustration of a typical workflow of data science agents.}
\label{fig:dsagent-workflow}
\end{figure}

The tools mentioned are part of an emerging category of software known as \textit{data science agents}, capable of executing a wide array of data-centric tasks, including manipulation, aggregation, visualization, and analysis, through natural language commands.
These agents primarily utilize LLMs to produce and implement code within designated data science platforms, such as Excel. 
Essential to their operation is the ability to comprehend the context of data and files in the ongoing session, along with the capability to verify and amend outputs as necessary, as discussed in studies \cite{cheng2023gpt,zhang2023data,tu2023should,chen2023teaching}. 
\autoref{fig:dsagent-workflow} depicts the typical workflow of a data science agent, highlighting its interactions with various components.

\begin{figure}[t]
\includegraphics[width=\linewidth,page=1]{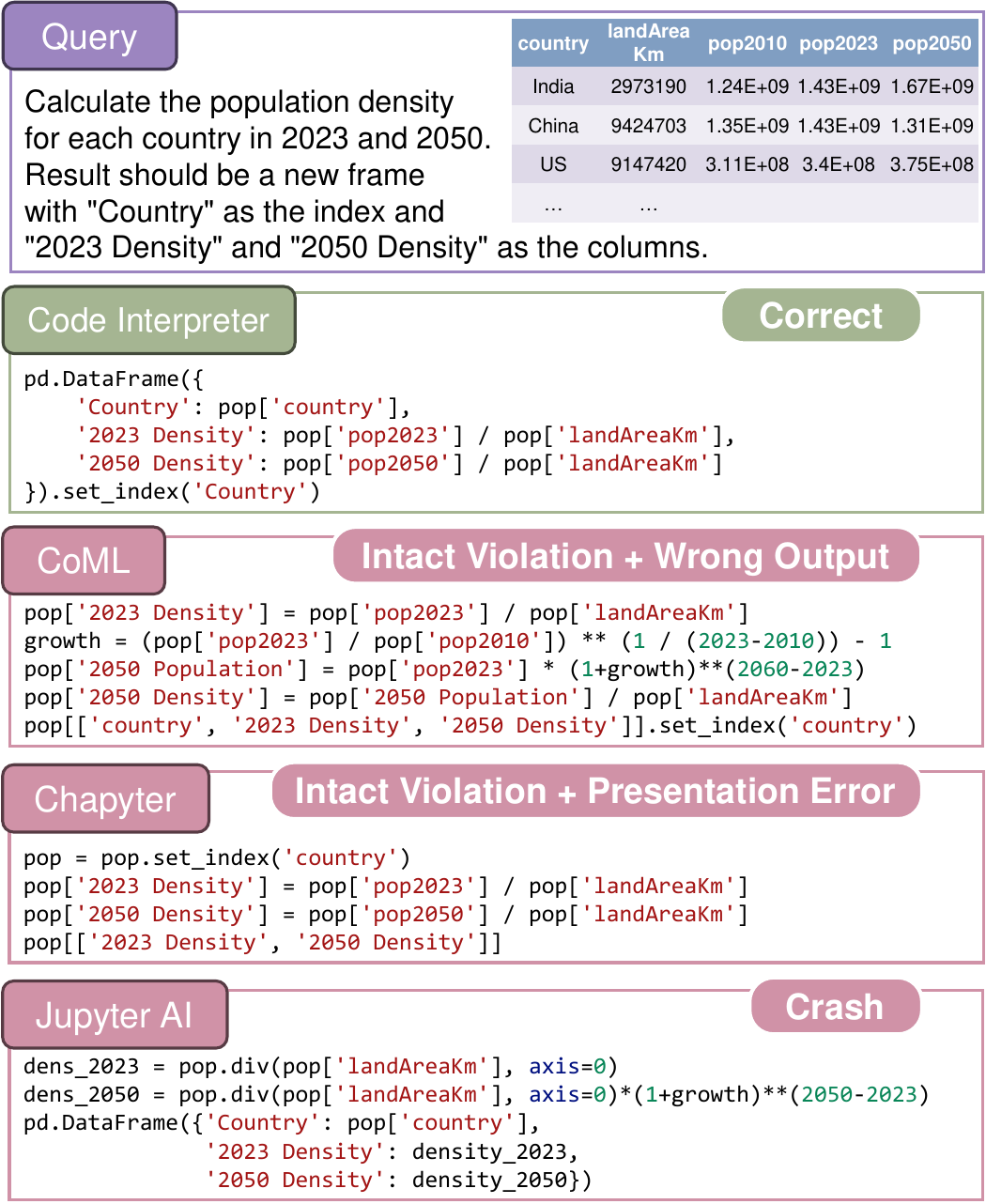}
\caption{A sample problem to test current data science agents. Code Interpreter is the only agent that produces the correct code to answer the query. CoML neglects the existing ``\texttt{pop2050}'' column in the table and predicts the population of 2050 on its own, which is not desired. Chapyter fails to capitalize the index ``\texttt{Country}'' and unintentionally modifies the data (``\texttt{pop}''), violating intactness. Jupyter AI divides strings by integers and cannot automatically recover from such failures.
}
\label{fig:dseval-intro}
\end{figure}

However, the reliability and accuracy of current data science agents can be inconsistent due to practical complexity of data science.
For instance, when we subjected four different agents to the same query, as shown in \autoref{fig:dseval-intro}, only one provided the correct response. 
The errors made by the others ranged from overlooking a data frame column, misinterpreting data types, failing to adhere to specified output formats, to altering the original data. 
These discrepancies can stem from various issues, including LLM limitations, unclear or inaccessible context, or a lack of failure recovery mechanisms.
Such shortcomings underscore the urgent need for focused research and enhancement of data science agents, with a particular emphasis on rigorous evaluation and benchmarking.

Evaluating data science agents is essential to pinpoint their capabilities and limitations, thereby informing future research trajectories. 
Yet, existing evaluation methodologies fall short of adequately addressing this need, being either insufficient or ill-suited for the task at hand.
Some existing works~\cite{zan2022cert,lai2023ds} only deliver incomplete evaluations of simple code completion or in-filling capability of LLMs, neglecting the whole problem-solving ability of agents.
Other recent works~\cite{cheng2023gpt,dibia2023lida} perform evaluations either on a limited scale or in a biased manner, mainly due to the heavy human efforts for dataset construction and agent evaluation.

In this paper, we introduce a novel benchmarking framework designed specifically for evaluations of data science agents.
Our contributions are three-fold.
First, we propose DSEval, an evaluation paradigm that enlarges the evaluation scope to the full lifecycle of LLM-based data science agents.
We also cover aspects including but not limited to the quality of the derived analytical solutions or machine learning models,
as well as potential side effects such as unintentional changes to the original data.
Second, we incorporate a novel bootstrapped annotation process letting LLM themselves generate and annotate the benchmarks with ``human in the loop''. 
A novel language (i.e., DSEAL) has been proposed and the derived four benchmarks have significantly improved the benchmark scalability and coverage, with largely reduced  human labor.
Third, based on DSEval and the four benchmarks, we conduct a comprehensive evaluation of various data science agents from different aspects.
Our findings reveal the common challenges and limitations of the current works, providing useful insights and shedding light on future research on LLM-based data science agents.

\section{Related Works}

\paragraph{Evaluating Code Generation Models.}
The field of LLMs \cite{brown2020language} has seen rapid progress, with many capable models that can generate high-quality natural language and codes for various domains and tasks~\cite{chen2021evaluating,roziere2023code}.
Benchmarks for these models \cite{chen2021evaluating} have also emerged.
Some of them are specifically designed for the data science domain, such as PandasEval / NumpyEval~\cite{zan2022cert}, DSP~\cite{chandel2022training} and DS-1000~\cite{lai2023ds}.
However, what these benchmarks provided were pre-written prompts, mainly for a fair comparison of completion of in-filling abilities of LLMs themselves.
They do not fully evaluate the skills of data science agents~\cite{zhang2023mlcopilot}, such as handling natural language interactions, managing runtime sessions, and assembling prompts.
Our evaluation scope is larger, which includes the full lifecycle of the agents.

\paragraph{Evaluating Agents.}
State-of-the-art LLMs~\cite{openai2023gpt4} have been used to power autonomous agents~\cite{autogpt,babyagi,wu2023autogen}, some of which get specialized in solving data science problems, such as data analysis, visualization and modeling~\cite{li2023camel,qian2023communicative,zhang2023mlcopilot}.
However, there is a lack of rigorous and systematic evaluation methods for these agents.
Some existing methods rely on huge human labor in problem collection and judgment, to assess the quality of the generated code or analysis~\cite{cheng2023gpt}, which incurs significant cost and restricts the scalability of benchmarks.
Some others resort to another more powerful LLM to score the output of the agent~\cite{dubois2023alpacafarm,dibia2023lida}, which may introduce bias and overlook errors.
Our work proposes a novel full-lifecycle evaluation paradigm to ensure robustness, and an additional LLM-bootstrapping annotation to enhance scalability and coverage.

\section{DSEval: Evaluation Paradigm for Data Science Agents}

To comprehensively and reliably evaluate a data science agent, we must first identify the evaluation scope, i.e., the ``lifecycle'' of an agent (\S~\ref{sec:agent-lifecycle}). Then we propose a paradigm that monitors the full lifecycle for complete assessments (\S~\ref{sec:agent-lifecycle-evaluation}).

\subsection{Evaluation Scope}
\label{sec:agent-lifecycle}

\begin{figure}[t]
\includegraphics[width=\linewidth,page=3]{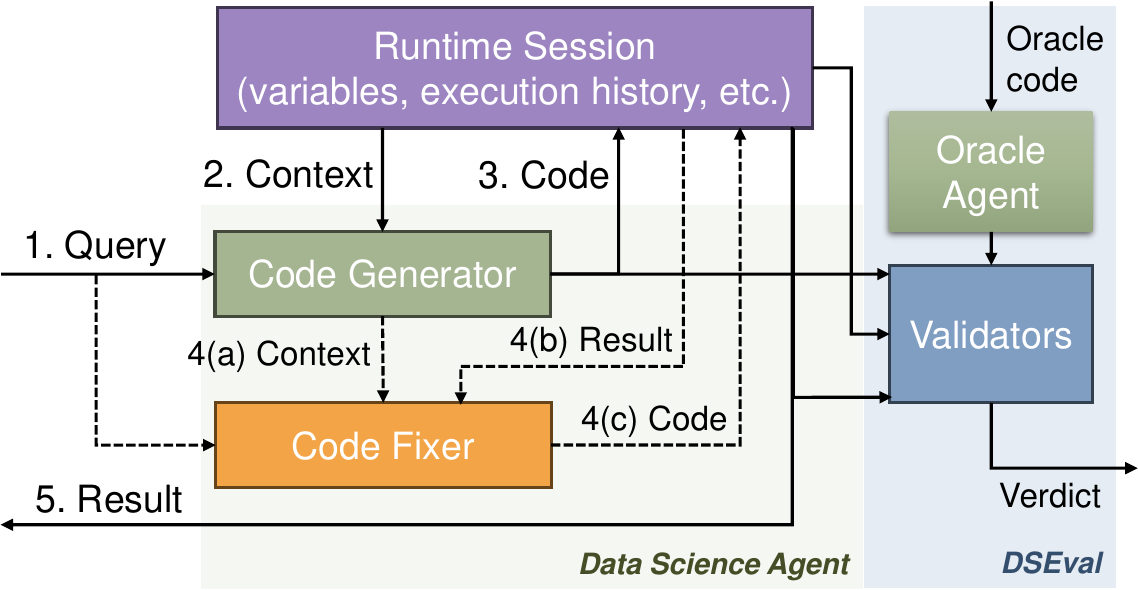}
\caption{Agent lifecycle, monitored by DSEval. Our evaluation scope is the green-shadowed area. DSEval monitors the full lifecycle.}
\label{fig:dseval-val-workflow}
\end{figure}

We argue that a robust data science agent depends not solely on the LLM capabilities, but also on the design of its other constituent components.
To identify the necessary scope for a comprehensive evaluation, we must first have a holistic perspective on the agent's lifecycle.

The lifecycle is depicted in the left part of \autoref{fig:dseval-val-workflow}.
First, the agent receives a ``\emph{query}'' (expressed in natural language).
Then it retrieves some additional contexts from a stateful ``\emph{runtime session}'', which is usually hosted by a data analysis platform (e.g., Jupyter), containing information like variables, execution history, files, etc.
A LLM-powered code generator then produces a code snippet based on the query and context.
The code is sent back to the runtime session for execution to get the result.
Optionally, a code fixer can help with error diagnosis and quality improvement (as done in tools like genai\footnote{\url{https://github.com/rgbkrk/genai}}).
The lifecycle can repeat itself for multi-rounds, with the runtime session keeping track of the conversation and execution history.

Our evaluation focuses on the holistic behavior of the data science agent, excluding implementation details of internal components such as code generators.
We design benchmarks with queries and runtime sessions as the only inputs, which essentially differs from existing code generation benchmarks~\cite{chen2021evaluating,zan2022cert}.

\subsection{Full-Lifecycle Monitoring}
\label{sec:agent-lifecycle-evaluation}

The holistic view of the agent lifecycle also makes us rethink the evaluation paradigm, and we conclude that ``\emph{every step and component involved in the lifecycle must be continuously monitored}''.
For instance, imagine a query requiring in-place dataset modifications. Here, validating the runtime session is crucial to confirm the accurate execution.
Hence, we design a validator module that is able to monitor the generated code, execution result, runtime session, etc.
Meanwhile, the validator leverages an oracle agent equipped with a reference code snippet, provided by benchmarks for comparison.
The process is illustrated in the right part of \autoref{fig:dseval-val-workflow}.

The validator implementations within the validator module are fully modular, with each implementation focusing on a specific phase (e.g., data matching with fuzzy order, or evaluating trained model performance on a held-out test dataset).
The full list and their usage frequencies are in \S~\ref{sec:benchmark-analysis} and \autoref{sec:validator-implementations}.
Notably, our focus is beyond correctness. For example, we implement an ``Intact'' validator, which tests whether the agent preserves the ``intactness'' of the session.
We implement this due to the belief that minimizing unintended changes is one important criterion of safety and reliability.

\section{Benchmarks based on DSEval}

Building upon the DSEval evaluation paradigm, we initiated the data collection and benchmark development process.
We came to realize that tremendous efforts were still required to properly rephrase queries, configure sessions, and adapt validators for each query.
Simple format conversion proved insufficient due to limitations in existing data sources: 
some data sources lack real-world complexity (e.g., pandas-exercises~\cite{pandasexercises}), while others address different-natured tasks (e.g., PandasEval~\cite{zan2022cert}).

To ensure the benchmark coverage with limited human efforts, we developed an ``LLM-bootstrapping annotation process'', leveraging LLMs to automatically create problemsets based on a minimal ``idea'', while incorporating human input.
This process is facilitated by the DSEAL (DSEval Annotation Language), which is designed to be compatible with the DSEval framework and easily comprehensible to LLMs.
In this section, we first introduce DSEAL (\S~\ref{sec:dseal}),
followed by a detailed description of the annotation process, including a Kaggle-inspired case study (\S~\ref{sec:annotation-process}).

\subsection{DSEAL: DSEval Annotation Language}
\label{sec:dseal}

DSEAL is essentially a language to describe ``problems''.
A \emph{problem} in DSEAL corresponds to one iteration depicted in \autoref{fig:dseval-val-workflow}, where a query is presented, and agents solve it and return results.
We define a ``\emph{problemset}'' as a sequence of interdependent problems, where later problems may have session or semantic dependencies on preceding ones.
A benchmark comprises multiple ``problemsets'', each of which is self-contained and isolated.  

The design of DSEAL is guided by three main objectives.
Firstly, it must be compatible with the DSEval framework, ensuring that its components are expressive enough to fit within the framework.
Secondly, it should be friendly to human annotators, for debuggability and ease of diagnosis.
Lastly, it needs to be easily understandable by LLMs to leverage their power for annotation purposes.

To achieve these goals, we have designed DSEAL as an extended version of the Python language.
Each problemset is represented as a Python (\texttt{*.py}) file, with problems separated by ``\texttt{\# \%\%}'' (cell syntax\footnote{\url{https://code.visualstudio.com/docs/datascience/jupyter-notebooks}}).
The code for oracle agents is written in Python, enabling direct execution and debugging using standard Python SDK.
We use triple-quoted strings with YAML syntax inside to ``configure'' the problem, including the query, validator configurations, execution restrictions, and external data required.
An example is provided in \autoref{fig:validation-language-example}.

\begin{figure}[t]
\begin{tcolorbox}
\scriptsize
\linespread{0.8}
\begin{verbatim}
# Previous problems...

# %%
"""
query: |
  Show the correlation between population
  density in 2023 and 2050, rounded to 2 decimals.
validator:
  template: basic
  namespace_intact:
    update: [pop]
  or:
    result:
      atol: 0
    output:
execution:
  forbid_names:
  - pop_heldout_test
  max_time: 0.5
data:
  pop.csv: https://.../pop.csv
"""
(pop['pop2023'] / pop['landAreaKm'])
  .corr(pop['pop2050'] / pop['landAreaKm']).round(2)

# Next problems...
\end{verbatim}
\end{tcolorbox}
\vspace{-3mm}
\caption{An example problemset written in DSEAL (DSEval Annotation Language).}
\vspace{1mm}
\label{fig:validation-language-example}
\end{figure}

\subsection{LLM-Bootstrapping Annotation Process}
\label{sec:annotation-process}

\begin{figure}[t]
\includegraphics[width=\linewidth,page=7]{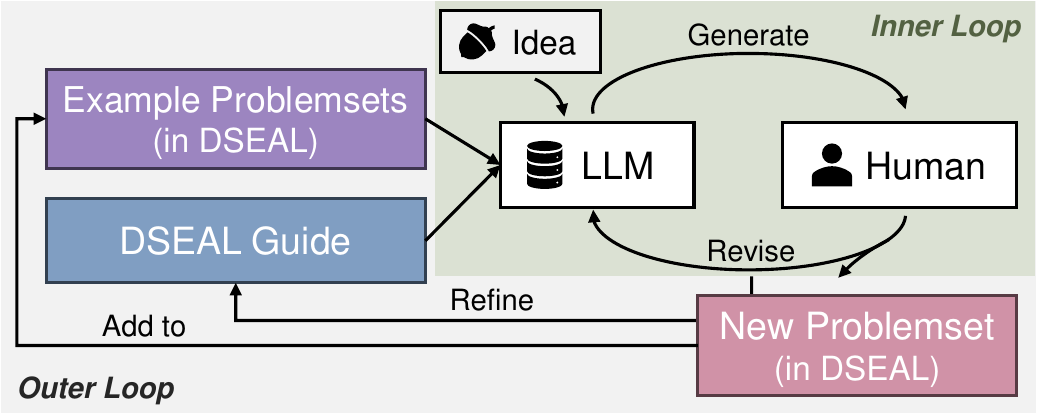}
\caption{Illustration of the LLM-Bootstrapping Annotation Process.}
\label{fig:bootstrapping-annotation-process}
\end{figure}

To alleviate human labor, we leverage the capability of LLMs to automatically annotate the benchmark as bootstrapping.
However, fully depending on LLMs may derive unreliable benchmarks even with state-of-the-art LLMs (further details are provided in the case study).
Therefore, we incorporate ``human-in-the-loop'' to further enhance the annotation.
The bootstrapping process involves an inner loop and an outer loop, as illustrated in \autoref{fig:bootstrapping-annotation-process}.

\paragraph{Inner Loop.}
To encourage LLMs to generate problemsets grounded in intended scenarios, we utilize ``idea seeds''.
These seeds anchor the generated problems to a specific scenario, promoting practicality and diversity across different outputs.
Additionally, we prompt LLMs with a ``guide'' containing instructions to format the problemset with DSEAL and ensure clarity and challenge. Few-shot examples from existing problemsets further enhance quality~\cite{kaplan2020scaling}.

Following the LLM's initial ``bootstrapping'' of a draft problemset, human experts step in for revision. Their focus lies in assessing clarity, diversity, and difficulty, and introducing necessary adjustments the LLM may struggle with independently. These adjustments can be partial, paving the way for the LLM to refine or enrich the problem set in an iterative loop.

\paragraph{Outer Loop.}
Once humans determine that no further adjustments are needed, the problemset is incorporated into the benchmark and serves as another example problemset of the LLM.
Additionally, revision comments are leveraged to enhance the DSEAL guide, preventing similar issues in future.
This loop, culminating in the accumulation of high-quality problem sets, exemplifies another form of ``bootstrapping'' within our process.

\begin{figure}[t]
\includegraphics[width=\linewidth,page=5]{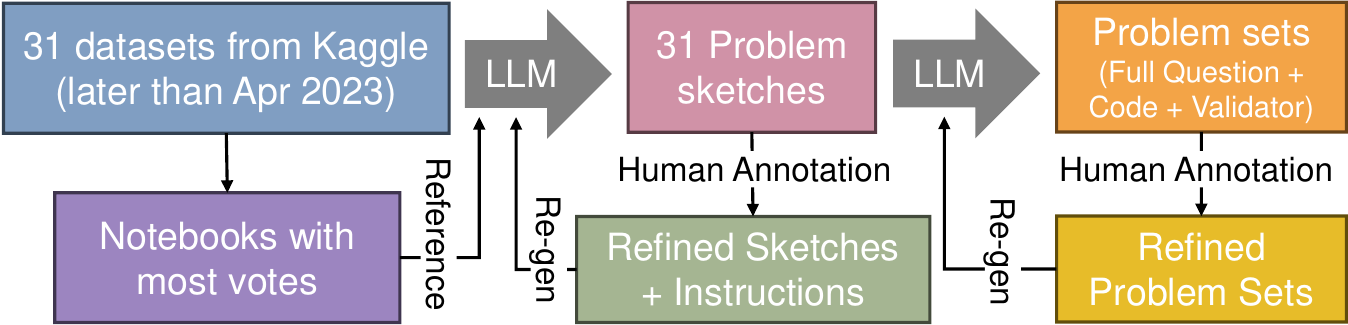}
\caption{Illustration of the annotation process on DSEval-Kaggle.}
\label{fig:kaggle-annotation-process}
\end{figure}

\paragraph{Case Study with DSEval-Kaggle.}
We selected 31 datasets published after April 2023, with data sizes less than 10 megabytes and more than 100 votes (by Sept. 2023).
We attached the most-voted\footnote{\url{https://www.kaggle.com/code?sortBy=voteCount}}
notebook associated with each dataset.
These 31 datasets and notebooks serve as the ``idea seeds''.

The inner loop has two primary stages in this case.
First, the targeted knowledge points and problemset sketches are created based on the dataset and notebook descriptions.
Second, the full problemset is generated based on the sketch from the first stage.
GPT-4 was used throughout the entire process.
The illustration is in \autoref{fig:kaggle-annotation-process}.

In early experiments, we encountered the following issues when relying solely on GPT-4 to generate the problemset.
(i) Lack of diversity due to repetitive generation results; for example, hypothesis test related queries appeared frequently.
(ii) Deviation from the actual dataset content, neglecting crucial initial steps like data cleaning.
(iii) Ambiguous queries resulting in vague or impossible-to-answer problems.
(iv) Incorrect solutions or incorrect validator configurations.
Interestingly, the first two issues can be effectively mitigated as the outer loop repeats.
The other two require resolution within the inner loop (i.e., from human revisions).

We ensure that all problems are revised at least once by human annotators, thus guaranteeing the quality of the benchmark. The entire annotation process required approximately 2.32 million prompts and 187k completion tokens on GPT-4, as well as 20 human hours.
We estimate a 3x reduction in human effort compared to purely manual methods like DS-1000~\cite{lai2023ds}.


\section{Statistics and Coverage}
\label{sec:benchmark-analysis}

Based on DSEval, we employed the annotation process to construct four benchmarks, detailed in \autoref{tab:dseval-subbenchmarks}.
These benchmarks encompass problem sets with diverse properties, ranging from straightforward tasks to more intricate challenges.
More technical details about how we created those benchmarks are available in \autoref{sec:benchmark-annotation-details}.

\paragraph{Validator Usages.}
Our evaluation process encompasses the entire lifecycle of data science agents.
We employ a total of nine validators, each targeting distinct facets within the lifecycle.
Details regarding their utilization are documented in Section~\ref{sec:validator-implementations}.
Within our benchmarks, data science agents undergo validation through 6.5 validators per problem on average.

\begin{table}[t]
\resizebox{\linewidth}{!}{%
\begin{tabular}{cccccc}
\toprule
Benchmark & \# Sets & \# Problems & Conversational & Realistic & Difficulty \\
\midrule
DSEval-Exercise & 21 & 187 & \Checkmark & \XSolidBrush & 17.3 \\
DSEval-SO & 202 & 202 & \XSolidBrush & \Checkmark & 16.1 \\
DSEval-LeetCode & 40 & 40 & \XSolidBrush & \XSolidBrush & 56.0 \\
DSEval-Kaggle & 31 & 396 & \Checkmark & \Checkmark & 35.9 \\
\bottomrule
\end{tabular}%
}
\caption{Overview of the four benchmarks.}
\label{tab:dseval-subbenchmarks}
\end{table}

\begin{figure}[t]
    \centering
    \includegraphics[width=\linewidth]{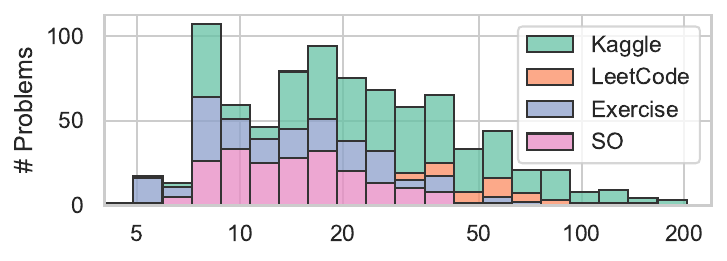}
    \caption{Difficulty distribution of the 4 benchmarks.}
    \label{fig:difficulty-distribution}
\end{figure}

\paragraph{Problem difficulty.}
For a better understanding of the performance across different difficulty levels, similar to previous studies~\cite{yu2018spider}, we quantify code complexity by considering the number of function calls, expressions, conditions, and loops in the reference code for each problem.
The distribution of problem difficulties is depicted in \autoref{fig:difficulty-distribution}, with the average difficulty detailed in \autoref{tab:dseval-subbenchmarks}.
We observe that DSEval-LeetCode poses the highest level of difficulty, while DSEval-Kaggle exhibits the most diverse range of difficulty levels.

\begin{figure}[t]
    \centering
    \begin{subfigure}[b]{.68\linewidth}
        \centering
        \includegraphics[width=\linewidth]{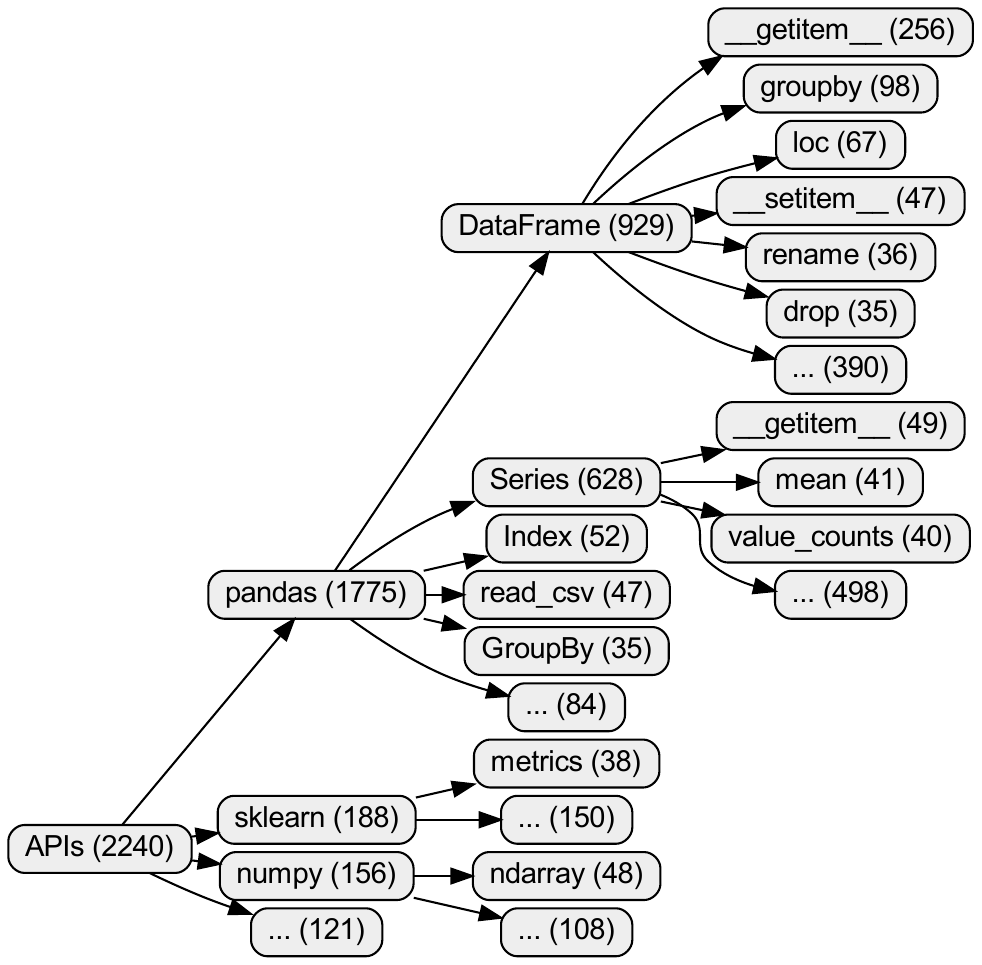}
        \caption{Data science APIs involved in the problems. In the parenthesis are the number of appearances.}
        \label{fig:api-tree}
    \end{subfigure}
    \hfill
    \begin{subfigure}[b]{0.28\linewidth}
        \centering
        \includegraphics[width=\linewidth]{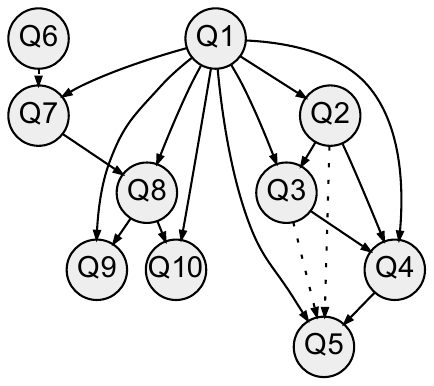}
        \caption{Dependency graph of problemset ``\emph{student-performance}'' in DSEval-Kaggle. Solid lines for ``session dependencies'' and dashed lines for ``semantic dependencies''.}
        \label{fig:dependency-graph}
    \end{subfigure}
    \caption{API coverage and dependency graph example.}

\end{figure}

\paragraph{API coverage.}
Collectively, the four benchmarks covered 2240 API calls spanning 448 distinct APIs within the oracle code.
These APIs are visualized in \autoref{fig:api-tree}.
Unsurprisingly, the most frequently utilized libraries are pandas, sklearn, and numpy.
In total, 12 libraries are covered, with imblearn, nltk, statsmodels, and catboost being the least frequently employed.
The most commonly occurring API is the \texttt{[]} operation of DataFrame, utilized for selecting indexes or columns.

\begin{figure}[t]
\includegraphics[width=\linewidth]{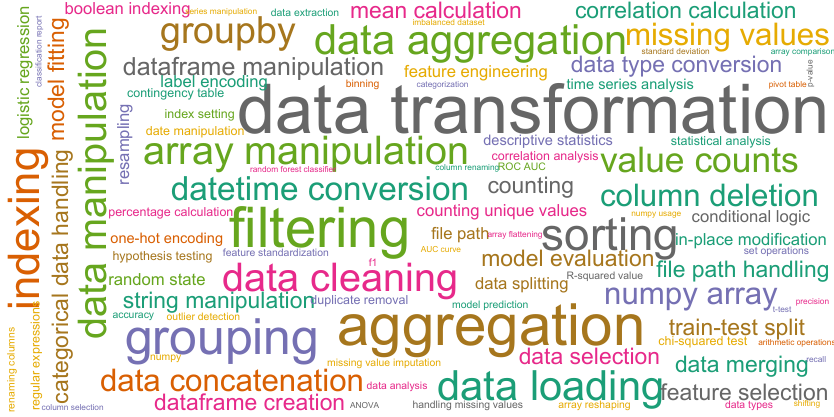}
\caption{Knowledge points involved in the problems.}
\label{fig:knowledge-wordcloud}
\end{figure}

\paragraph{Knowledge points coverage.}
We use GPT-3.5 to summarize the data science knowledge points essential for solving each problem. As illustrated in the word cloud of \autoref{fig:knowledge-wordcloud}, the benchmarks focus on fundamental data processing concepts such as data transformation, aggregation, filtering, sorting, and grouping, as well as encompassing machine learning concepts like outlier detection, imbalanced dataset handling, and feature selection.

\paragraph{Problem dependencies.}
DSEval-Kaggle and DSEval-Exercise are two conversational benchmarks where there could be interdependences among problems. 
We define ``session dependency'' as a scenario where a variable from a previous problem is used in a subsequent problem, and ``semantic dependency'' as a situation where the comprehension of a later query relies on the context of a preceding query. 
We visualize dependency graphs for each problem set (see \autoref{fig:dependency-graph} for example). 
On average, we observe an in-degree of 2.08 across all graphs. Regarding the maximum dependency chain length, the longest chain spans 10 dependencies, with an average chain length of 4.06.

\paragraph{Session contexts.}
A major challenge in our proposed benchmark lies in retrieving and representing contexts from runtime sessions.
On average, we estimate that each problem involves 3.68 variables, with a maximum of up to 29.
The total data size of these variables is 1.12 kilobytes at the median, and can reach up to 268 megabytes in extreme cases.

\section{Evaluation}

\subsection{Setups}

\begin{table*}[t]
\centering
\resizebox{\textwidth}{!}{\begin{tabular}{c|cccc|cccc|ccc|ccc}
\toprule
\multirow{2}*{Framework} & \multicolumn{4}{c|}{DSEval-Kaggle} & \multicolumn{4}{c|}{DSEval-Exercise} & \multicolumn{3}{c|}{DSEval-LeetCode} & \multicolumn{3}{c}{DSEval-SO} \\
& \begin{tabular}{@{}c@{}}Pass \\ Rate\end{tabular} & \begin{tabular}{@{}c@{}}Error \\ Prop\end{tabular} & \begin{tabular}{@{}c@{}}w/o \\ Intact\end{tabular} & \begin{tabular}{@{}c@{}}w/o \\ PE\end{tabular}  & \begin{tabular}{@{}c@{}}Pass \\ Rate\end{tabular} & \begin{tabular}{@{}c@{}}Error \\ Prop\end{tabular} & \begin{tabular}{@{}c@{}}w/o \\ Intact\end{tabular} & \begin{tabular}{@{}c@{}}w/o \\ PE\end{tabular}  & \begin{tabular}{@{}c@{}}Pass \\ Rate\end{tabular} & \begin{tabular}{@{}c@{}}w/o \\ Intact\end{tabular} & \begin{tabular}{@{}c@{}}w/o \\ PE\end{tabular}  & \begin{tabular}{@{}c@{}}Pass \\ Rate\end{tabular} & \begin{tabular}{@{}c@{}}w/o \\ Intact\end{tabular} & \begin{tabular}{@{}c@{}}w/o \\ PE\end{tabular} \\
\midrule
Chapyter~\cite{chapyter} & {\cellcolor[HTML]{E0D6E3}} \color[HTML]{000000} 34.1 & {\cellcolor[HTML]{F1F0F1}} \color[HTML]{000000} 26.0 & {\cellcolor[HTML]{DDD1E0}} \color[HTML]{000000} 35.6 & {\cellcolor[HTML]{B390BB}} \color[HTML]{F1F1F1} 55.3  & {\cellcolor[HTML]{E1D7E3}} \color[HTML]{000000} 39.6 & {\cellcolor[HTML]{F1F0F1}} \color[HTML]{000000} 28.3 & {\cellcolor[HTML]{DDD0DF}} \color[HTML]{000000} 42.2 & {\cellcolor[HTML]{B28FBA}} \color[HTML]{F1F1F1} 70.6  & {\cellcolor[HTML]{D7C7DA}} \color[HTML]{000000} 45.0 & {\cellcolor[HTML]{D7C7DA}} \color[HTML]{000000} 45.0 & {\cellcolor[HTML]{B796BE}} \color[HTML]{F1F1F1} 60.0  & {\cellcolor[HTML]{F1F0F1}} \color[HTML]{000000} 46.5 & {\cellcolor[HTML]{ECE9ED}} \color[HTML]{000000} 48.5 & {\cellcolor[HTML]{D1BFD5}} \color[HTML]{000000} 59.9 \\
ChatDev~\cite{qian2023communicative} & {\cellcolor[HTML]{000000}} \color[HTML]{F1F1F1} - & {\cellcolor[HTML]{000000}} \color[HTML]{F1F1F1} - & {\cellcolor[HTML]{000000}} \color[HTML]{F1F1F1} - & {\cellcolor[HTML]{000000}} \color[HTML]{F1F1F1} -  & {\cellcolor[HTML]{000000}} \color[HTML]{F1F1F1} - & {\cellcolor[HTML]{000000}} \color[HTML]{F1F1F1} - & {\cellcolor[HTML]{000000}} \color[HTML]{F1F1F1} - & {\cellcolor[HTML]{000000}} \color[HTML]{F1F1F1} -  & {\cellcolor[HTML]{F1F0F1}} \color[HTML]{000000} 32.5 & {\cellcolor[HTML]{F1F0F1}} \color[HTML]{000000} 32.5 & {\cellcolor[HTML]{CCB7D1}} \color[HTML]{000000} 50.0  & {\cellcolor[HTML]{F1F0F1}} \color[HTML]{000000} 35.1 & {\cellcolor[HTML]{F1F0F1}} \color[HTML]{000000} 35.1 & {\cellcolor[HTML]{EDEAED}} \color[HTML]{000000} 37.6 \\
CoML~\cite{zhang2023mlcopilot} & {\cellcolor[HTML]{AA81B3}} \color[HTML]{F1F1F1} 59.8 & {\cellcolor[HTML]{B08BB8}} \color[HTML]{F1F1F1} 56.8 & {\cellcolor[HTML]{A77DB1}} \color[HTML]{F1F1F1} 61.1 & {\cellcolor[HTML]{A275AC}} \color[HTML]{F1F1F1} 63.6  & {\cellcolor[HTML]{A77CB0}} \color[HTML]{F1F1F1} 78.6 & {\cellcolor[HTML]{A77CB0}} \color[HTML]{F1F1F1} 78.6 & {\cellcolor[HTML]{A67BAF}} \color[HTML]{F1F1F1} 79.1 & {\cellcolor[HTML]{A376AD}} \color[HTML]{F1F1F1} 81.3  & {\cellcolor[HTML]{DCCFDF}} \color[HTML]{000000} 42.5 & {\cellcolor[HTML]{DCCFDF}} \color[HTML]{000000} 42.5 & {\cellcolor[HTML]{B28EBA}} \color[HTML]{F1F1F1} 62.5  & {\cellcolor[HTML]{A57AAF}} \color[HTML]{F1F1F1} 78.2 & {\cellcolor[HTML]{A275AC}} \color[HTML]{F1F1F1} 79.7 & {\cellcolor[HTML]{A275AC}} \color[HTML]{F1F1F1} 79.7 \\
Code Interpreter API~\cite{codeinterpreterapi} & {\cellcolor[HTML]{CFBBD3}} \color[HTML]{000000} 42.4 & {\cellcolor[HTML]{D0BDD4}} \color[HTML]{000000} 41.7 & {\cellcolor[HTML]{CCB6D0}} \color[HTML]{000000} 43.9 & {\cellcolor[HTML]{C5ACCB}} \color[HTML]{000000} 47.0  & {\cellcolor[HTML]{B796BF}} \color[HTML]{F1F1F1} 67.4 & {\cellcolor[HTML]{B695BE}} \color[HTML]{F1F1F1} 67.9 & {\cellcolor[HTML]{B694BD}} \color[HTML]{F1F1F1} 68.4 & {\cellcolor[HTML]{B18CB9}} \color[HTML]{F1F1F1} 71.7  & {\cellcolor[HTML]{D7C7DA}} \color[HTML]{000000} 45.0 & {\cellcolor[HTML]{D7C7DA}} \color[HTML]{000000} 45.0 & {\cellcolor[HTML]{C2A6C8}} \color[HTML]{000000} 55.0  & {\cellcolor[HTML]{D5C4D9}} \color[HTML]{000000} 58.4 & {\cellcolor[HTML]{BD9FC4}} \color[HTML]{F1F1F1} 68.3 & {\cellcolor[HTML]{C3A8C9}} \color[HTML]{000000} 65.8 \\
Jupyter-AI~\cite{jupyterai} & {\cellcolor[HTML]{BB9CC2}} \color[HTML]{F1F1F1} 51.8 & {\cellcolor[HTML]{D7C8DA}} \color[HTML]{000000} 38.4 & {\cellcolor[HTML]{B998C0}} \color[HTML]{F1F1F1} 52.8 & {\cellcolor[HTML]{AD87B6}} \color[HTML]{F1F1F1} 58.1  & {\cellcolor[HTML]{A77CB0}} \color[HTML]{F1F1F1} 78.6 & {\cellcolor[HTML]{CAB3CF}} \color[HTML]{000000} 55.1 & {\cellcolor[HTML]{A67BAF}} \color[HTML]{F1F1F1} 79.1 & {\cellcolor[HTML]{A275AC}} \color[HTML]{F1F1F1} 81.8  & {\cellcolor[HTML]{BC9EC3}} \color[HTML]{F1F1F1} 57.5 & {\cellcolor[HTML]{BC9EC3}} \color[HTML]{F1F1F1} 57.5 & {\cellcolor[HTML]{A275AC}} \color[HTML]{F1F1F1} 70.0  & {\cellcolor[HTML]{E9E4EA}} \color[HTML]{000000} 50.0 & {\cellcolor[HTML]{E9E4EA}} \color[HTML]{000000} 50.0 & {\cellcolor[HTML]{DACCDD}} \color[HTML]{000000} 56.4 \\
\bottomrule
\end{tabular}%
}
\caption{Performance of agent frameworks on DSEval benchmarks. We compare: pass rate, pass rate with error propagation, pass rate without the constraint of intact violation, and pass rate without considering presentation error. ChatDev is only evaluated on DSEval-LeetCode and DSEval-SO due to the difficulty of injecting complex context.}
\label{tab:compare-agent-frameworks}
\end{table*}

\paragraph{Error Categories.}
When an agent fails to successfully respond to a problem, the errors in an agent-generated code snippet can be classified into eight major categories, which can be further broken down into 32 subcategories.
The complete catalog is presented in \autoref{fig:verdict-pie-chart} and \autoref{sec:appendix-verdict-catalog}. Two common errors are highlighted below:

\begin{itemize}
\item \emph{Presentation Error:} This occurs when the result is almost correct but problematic in terms of format or presentation approach. For example, the agent might fail to capitalize a column name as instructed or erroneously print results to the console instead of placing them in cell outputs.

\item \emph{Intact Violation:} Happens when the solution is almost correct except for violating the concept of intactness. This typically occurs when the computation requires some intermediate columns and the agent modifies the original data, which is unnecessary.
\end{itemize}

\paragraph{Metrics.}
The ``\emph{Pass Rate}'', which is the number of problems passed divided by all problems in the benchmark, is the default metric used to assess the quality of an agent.
By default, the runtime session is set to the ground-truth state before evaluating each problem.
We refer to ``\emph{error propagation}'' as a special setting where erroneous states accumulate to affect future problems within the same problem set.
Additionally, we compute the pass rate while ignoring intact violations and presentation errors (``\emph{w/o Intact}'' and ``\emph{w/o PE}'' respectively), as they can be considered correct in a looser setting.

\subsection{Evaluating Data Science Agents}

\begin{figure}[t]
\centering
\includegraphics[width=\linewidth]{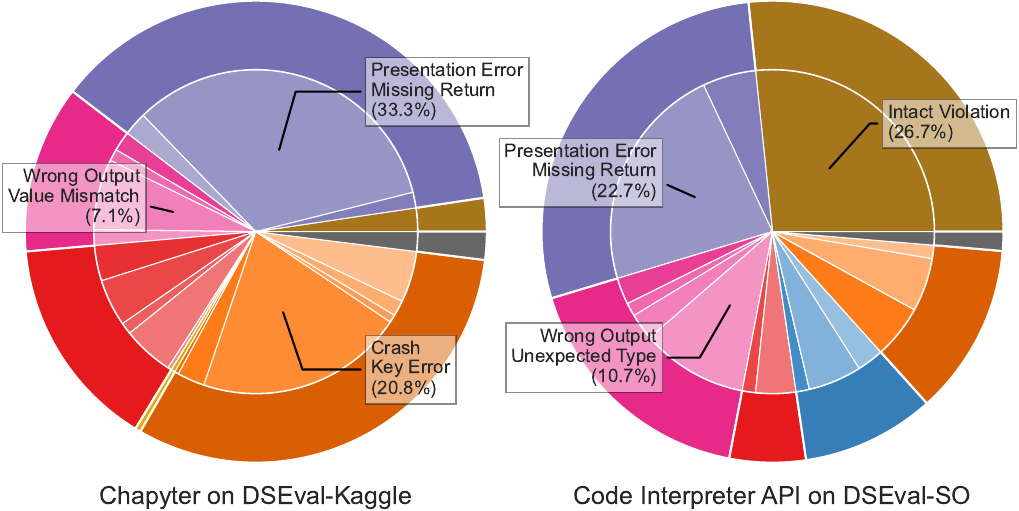}
\caption{Two examples of error type breakdowns. More in \autoref{fig:verdict-pie-chart-full}.}
\label{fig:verdict-pie-chart}
\end{figure}

We evaluate five popular LLM-based agents that are currently applicable to data science scenarios: Chapyter, ChatDev, CoML, Code Interpreter API, and Jupyter-AI (summarized in \S~\ref{sec:intro-benchmarked-agents}).
These selected agents cover mainstream agent-building approaches, including function calls, expert knowledge, and multi-agent communications.
For fair comparisons, we use GPT-3.5 (v1106) with a temperature of 0 as backend LLMs for all agents.

The key observations from \autoref{tab:compare-agent-frameworks} are as follows:
(i) Chapyter is the worst-performing agent, but its pass rate significantly improves when presentation errors are ignored.
(ii) CoML is the best for most benchmarks, except for LeetCode, where Jupyter-AI outperforms greatly.
(iii) When errors propagate, Chapyter and Jupyter-AI suffer greatly, yet the other two frameworks remain stable.
(iv) Intact violations sometimes occur, but not frequently.

To gain a better understanding of these error types, we did several case studies and visualized the percentages of error causes in \autoref{tab:compare-agent-frameworks} in \autoref{fig:verdict-pie-chart}. We can see that the primary issue with Chapyter is missing returns (e.g., using ``\texttt{print()}'' instead of ``\texttt{return}'' to show the output) and key errors (e.g., referencing non-existing columns). Code Interpreter API on DSEval-SO often triggers intact violations, as the framework has a tendency to perform inplace modifications to existing variables.

\subsection{Context Selection and Representations}

A key distinction among the agent frameworks lies in how they select and represent contexts from the sessions.
Contexts are crucial for agents as they complement the missing parts of the query.
Under the scope of our benchmarks, contexts are roughly categorized into variable descriptions and executed code history.

We conduct experiments with different combinations and orders of variable descriptions, code histories, and queries.
We pick CoML as the baseline agent framework as it appears to be the best-performing one in previous experiments.
The results are shown in \autoref{tab:compare-context-orders}.
We observe that without any context, LLMs struggle to produce correct results.
Providing code history and variable descriptions as context improve performance of agents.
Code history seems to be more essential, especially for simpler tasks like DSEval-Exercise.
The order of the context also has a slight impact: placing variable descriptions and queries at the end of the input tends to improve the results.

\begin{table}[t]
\centering
\small
\resizebox{\linewidth}{!}{\begin{tabular}{c|cc|cc}
\toprule
\multirow{2}*{Context} & \multicolumn{2}{c}{DSEval-Kaggle} & \multicolumn{2}{c}{DSEval-Exercise} \\
 & Pass Rate & w/ Error Prop & Pass Rate & w/ Error Prop \\
\midrule
Q & 13.9 & 13.9 & 13.9 & 13.9 \\
C+Q & 53.8 & 40.4 & \bfseries 81.3 & \bfseries 80.7 \\
V+Q & 52.3 & 51.5 & 73.3 & 71.1 \\
C+V+Q & \bfseries 61.4 & 52.5 & 80.7 & 80.2 \\
V+C+Q & 59.8 & \bfseries 56.8 & 78.6 & 78.6 \\
Q+V+C & 58.3 & 53.5 & 74.3 & 71.7 \\
\bottomrule
\end{tabular}%
}
\caption{Comparison of combinations in the context. ``C'' stands for ``Code history'', ``V'' stands for ``Variable descriptions'' and ``Q'' stands for ``Query''.}
\label{tab:compare-context-orders}
\end{table}

\begin{figure}[t]
\centering
\includegraphics[width=\linewidth,page=6]{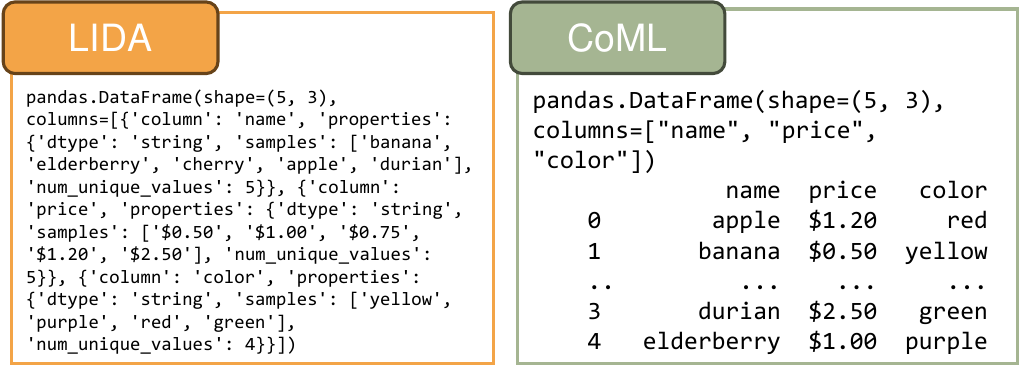}
\caption{Illustration of data table formatter in LIDA and CoML.}
\label{fig:data-table-formatter}
\end{figure}

Encoding the context into the prompt poses another challenge.
Previous work~\cite{sui2024table} has explored this issue and proposed different methods to compress megabytes of data into dozens of tokens.
We evaluate the approaches used in LIDA~\cite{dibia2023lida} and CoML, with differences shown in \autoref{fig:data-table-formatter}.

As shown in \autoref{tab:compare-representations}, LIDA and CoML have similar performance on DSEval-Kaggle, but LIDA outperforms CoML on DSEval-Exercise.
This difference in performance could be due to LIDA encoding more information such as the data type and the unique-value count of each column.
However, this also means that LIDA consumes more tokens than CoML to represent the same table.

\begin{table}[t]
\centering
\small
\resizebox{\linewidth}{!}{\begin{tabular}{c|cc|cc}
\toprule
\multirow{2}*{Format} & \multicolumn{2}{c}{DSEval-Kaggle} & \multicolumn{2}{c}{DSEval-Exercise} \\
& Pass Rate & \# Tokens & Pass Rate & \# Tokens \\
\midrule
CoML & 59.8 & 2963.7 & 78.6 & 2126.3 \\
LIDA & 59.8 & 4192.7 & 82.4 & 2547.6 \\
\bottomrule
\end{tabular}%
}
\caption{Comparison of pass rate and consumed prompt tokens for different code and data encodings in prompts.}
\label{tab:compare-representations}
\end{table}

\subsection{Evaluating LLMs}

\begin{figure}[t]
\centering
\includegraphics[width=\linewidth]{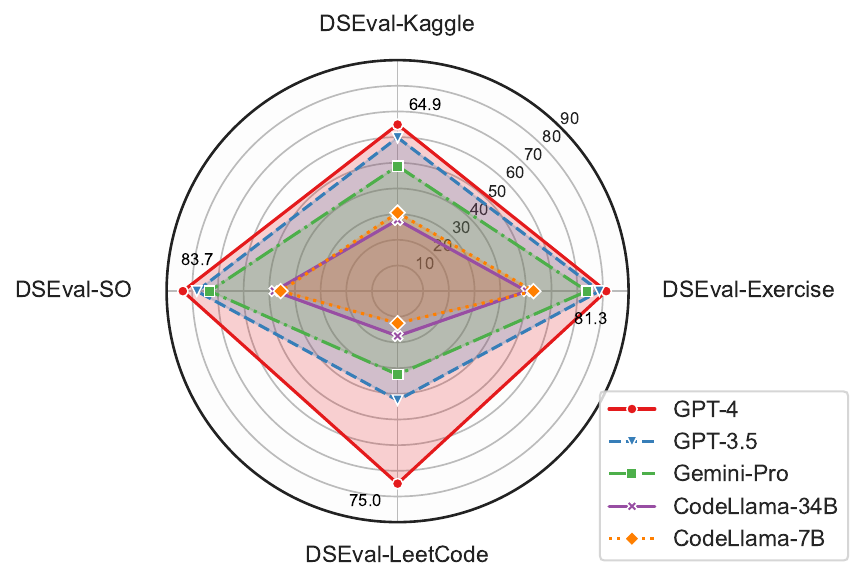}
\caption{Performance of CoML combined with different LLMs on four benchmarks of DSEval.}
\label{fig:compare-models}
\end{figure}

We experimentally combine CoML with different LLMs and compare their performance.
The results are shown in \autoref{fig:compare-models}.
In addition to GPT-3.5, which we have already tried, we include four more models for comparison: GPT-4~\cite{openai2023gpt4}, Gemini-Pro~\cite{team2023gemini}, CodeLlama-7B~\cite{roziere2023code}, and CodeLlama-34B.
The rank of the models is approximately as follows: CodeLlama-7B $\approx$ CodeLlama-34B $<$ Gemini-Pro $<$ GPT-3.5 $<$ GPT-4. More details are in \S~\ref{sec:compare-models-more}.

\subsection{Self Repair}

To evaluate the diagnostic and self-repair abilities of data science agents, we apply the self-debug~\cite{chen2023teaching} to the DSEval benchmarks.
We use the CoML implementation, which sends the output and errors to LLMs for line-by-line analysis and feedback, before receiving a revised code.
We do not use any hints from validators during this process.
It repeats until we obtain a correct result or reach the maximum number of attempts.
We also compare self-debug with a simple resampling baseline, which resamples a new code snippet if the previous one is incorrect.

\begin{figure}[t]
\centering
\includegraphics[width=\linewidth]{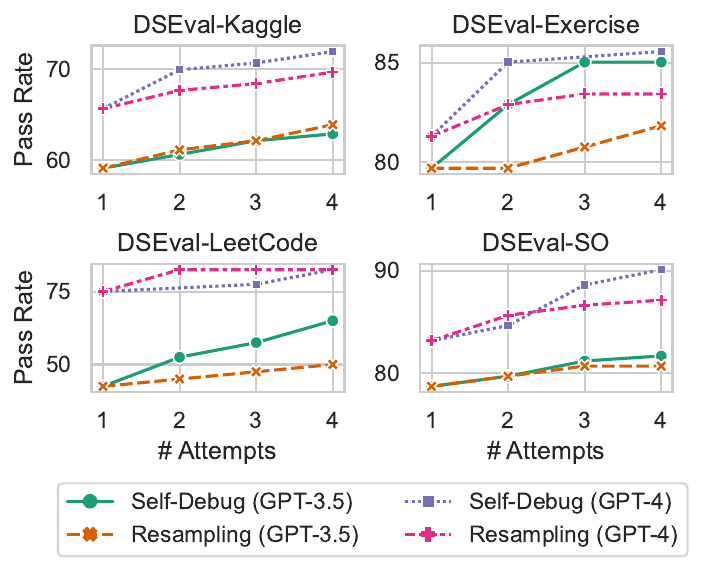}
\caption{Self-debug versus vanilla resampling.}
\label{fig:compare-retries}
\end{figure}

\autoref{fig:compare-retries} shows two main findings.
First, both self-debug and resampling enhance performance, but self-debug is generally more effective.
Second, models with lower capabilities (e.g., GPT-3.5) can outperform models with higher capabilities (e.g., GPT-4) with enough self-repair attempts.

We also analyzed the error types that can be fixed via self-repair on DSEval-Kaggle and found that around half of them are ``Crash'' errors.
Among all the ``Crash'' errors, 15\% will still crash after the 4th attempt, and 41\% will turn into other error types.
Among all error types except "Crash", the type that is most likely to be fixed is ``Presentation Error'', with a fixed probability of 20\% (4 / 20).
This suggests there is room for improvement in current self-repairing techniques.

\section{Conclusion}

In this paper, we introduce DSEval, an evaluation paradigm for data science agents.
Based on DSEval, we created 4 benchmarks that cover different aspects of data science tasks, and existing agents were evaluated and analyzed on the benchmarks.
We aim to establish a standard for evaluating data science agents and we welcome more contributions of problemsets from the community.

\section{Ethical Considerations}

Modern data science agents have made it easier to analyze, visualize and process data.
However, such agents can also pose serious risks if they are not used carefully.
For example, a data science agent can alter the data without the user's awareness, or generate a misleading data analysis that appears to be correct but is actually erroneous.

Our work is the first to address these issues in a comprehensive way.
For instance, we developed a validator that can track the full lifecycle of agent and assess whether the agent causes any unwanted changes (via ``Intact'' validator).
We think future data science agents should follow our benchmarks as a reference, to ensure that they produce reliable and safe outcomes.

\section{Limitations}

\paragraph{Evaluating Planning Ability.}
The goal of planning is to break down a complex task into several simple, executable tasks, which is a key skill of LLM agents~\cite{shen2023hugginggpt,wu2023autogen}.
In this paper, we focus on evaluating data science agents' performance on single tasks.
Although some tasks (especially those in DSEval-Kaggle) are very complex and require careful planning to solve, we did not include high-level data science tasks that are vague and open-ended, such as ``design a data pipeline that will win this Kaggle competition''.
However, we think that DSEval framework can also support those tasks, as long as the evaluation criteria (i.e., validator) are properly defined and configured.

\paragraph{Reproducibility and Stableness.}
We conducted extensive evaluations and obtained some interesting insights, but unfortunately we could not repeat every experiment to check the reproducibility of each result due to the budget constraint.
Instead, we focused more on evaluating different settings and benchmarks, which we believe are more informative.
In \autoref{tab:stableness}, we verified some of the experiments by either repeating them, using a different model version, or changing a parameter (e.g., the temperature).
We observed that the results are not very stable and can vary by up to $\pm 2\%$.
On DSEval-LeetCode, the variation is even more significant, probably because the benchmark only has 40 problems.
However, we remark that we have published 4 benchmarks based on DSEval, multiple results on different benchmarks can still have some significance.
We encourage the community to take the average of multiple runs when possible.

\begin{table}[t]
\centering
\small
\resizebox{\linewidth}{!}{%
\begin{tabular}{ccccccc}
\toprule
Model & Temp & Repeat & Kaggle & Exercise & LeetCode & SO \\
\midrule
CodeLlama-7B & 0.0 & 0 & 30.6 & 52.9 & 12.5 & 45.5 \\
CodeLlama-7B & 0.0 & 1 & 30.3 & 52.9 & 12.5 & 45.5 \\
CodeLlama-7B & 0.5 &  & 30.8 & 46.0 & 15.0 & 47.0 \\
\midrule
CodeLlama-34B & 0.0 & 0 & 27.8 & 50.3 & 17.5 & 48.0 \\
CodeLlama-34B & 0.0 & 1 & 27.8 & 49.7 & 10.0 & 48.0 \\
CodeLlama-34B & 0.5 &  & 25.5 & 48.1 & 30.0 & 45.5 \\
\midrule
Gemini-Pro & 0.0 & 0 & 48.7 & 73.8 & 32.5 & 73.3 \\
Gemini-Pro & 0.0 & 1 & 47.2 & 73.3 & 32.5 & 73.3 \\
Gemini-Pro & 0.5 &  & 43.4 & 65.2 & 37.5 & 66.8 \\
\midrule
GPT-3.5 & 0.0 & 0 & 59.8 & 78.6 & 42.5 & 80.2 \\
GPT-3.5 & 0.0 & 1 & 60.6 & 80.7 & 45.0 & 79.2 \\
GPT-3.5 & 0.0 & 2 & 60.4 & 79.7 & 42.5 & 78.2 \\
GPT-3.5 & 0.5 &  & 58.8 & 79.1 & 47.5 & 80.7 \\
GPT-3.5 (v0613) & 0.0 &  & 61.9 & 80.7 & 37.5 & 79.2 \\
\midrule
GPT-4 & 0.0 & 0 & 64.9 & 81.3 & 75.0 & 83.7 \\
GPT-4 & 0.0 & 1 & 64.1 & 81.3 & 70.0 & 84.7 \\
GPT-4 & 0.0 & 2 & 64.6 & 82.4 & 77.5 & 85.1 \\
GPT-4 & 0.5 &  & 64.6 & 82.4 & 72.5 & 85.1 \\
GPT-4-32k & 0.0 &  & 65.4 & 80.2 & 67.5 & 80.7 \\
\bottomrule
\end{tabular}%
}
\caption{Reproducibility test by repeating the experiment, and possibly varying the temperature and model version. Default versions for GPT-3.5 and GPT-4 are v1106.}
\label{tab:stableness}
\end{table}

\bibliography{main}

\begin{thebibliography}{34}
\expandafter\ifx\csname natexlab\endcsname\relax\def\natexlab#1{#1}\fi

\bibitem[{Brown et~al.(2020)Brown, Mann, Ryder, Subbiah, Kaplan, Dhariwal,
  Neelakantan, Shyam, Sastry, Askell et~al.}]{brown2020language}
Tom Brown, Benjamin Mann, Nick Ryder, Melanie Subbiah, Jared~D Kaplan, Prafulla
  Dhariwal, Arvind Neelakantan, Pranav Shyam, Girish Sastry, Amanda Askell,
  et~al. 2020.
\newblock Language models are few-shot learners.
\newblock \emph{Advances in neural information processing systems},
  33:1877--1901.

\bibitem[{Chandel et~al.(2022)Chandel, Clement, Serrato, and
  Sundaresan}]{chandel2022training}
Shubham Chandel, Colin~B Clement, Guillermo Serrato, and Neel Sundaresan. 2022.
\newblock Training and evaluating a jupyter notebook data science assistant.
\newblock \emph{arXiv preprint arXiv:2201.12901}.

\bibitem[{chapyter(2023)}]{chapyter}
chapyter. 2023.
\newblock chapyter.
\newblock \url{https://github.com/chapyter/chapyter}.

\bibitem[{Chen et~al.(2021)Chen, Tworek, Jun, Yuan, Pinto, Kaplan, Edwards,
  Burda, Joseph, Brockman et~al.}]{chen2021evaluating}
Mark Chen, Jerry Tworek, Heewoo Jun, Qiming Yuan, Henrique Ponde de~Oliveira
  Pinto, Jared Kaplan, Harri Edwards, Yuri Burda, Nicholas Joseph, Greg
  Brockman, et~al. 2021.
\newblock Evaluating large language models trained on code.
\newblock \emph{arXiv preprint arXiv:2107.03374}.

\bibitem[{Chen et~al.(2023)Chen, Lin, Sch{\"a}rli, and Zhou}]{chen2023teaching}
Xinyun Chen, Maxwell Lin, Nathanael Sch{\"a}rli, and Denny Zhou. 2023.
\newblock Teaching large language models to self-debug.
\newblock \emph{arXiv preprint arXiv:2304.05128}.

\bibitem[{Cheng et~al.(2023)Cheng, Li, and Bing}]{cheng2023gpt}
Liying Cheng, Xingxuan Li, and Lidong Bing. 2023.
\newblock Is gpt-4 a good data analyst?
\newblock \emph{arXiv preprint arXiv:2305.15038}.

\bibitem[{Dibia(2023)}]{dibia2023lida}
Victor Dibia. 2023.
\newblock Lida: A tool for automatic generation of grammar-agnostic
  visualizations and infographics using large language models.
\newblock \emph{arXiv preprint arXiv:2303.02927}.

\bibitem[{Dubois et~al.(2023)Dubois, Li, Taori, Zhang, Gulrajani, Ba, Guestrin,
  Liang, and Hashimoto}]{dubois2023alpacafarm}
Yann Dubois, Xuechen Li, Rohan Taori, Tianyi Zhang, Ishaan Gulrajani, Jimmy Ba,
  Carlos Guestrin, Percy Liang, and Tatsunori~B. Hashimoto. 2023.
\newblock \href {http://arxiv.org/abs/2305.14387} {Alpacafarm: A simulation
  framework for methods that learn from human feedback}.

\bibitem[{guipsamora(2020)}]{pandasexercises}
guipsamora. 2020.
\newblock pandas\_exercises.
\newblock \url{https://github.com/guipsamora/pandas\_exercises}.

\bibitem[{jupyterlab(2023)}]{jupyterai}
jupyterlab. 2023.
\newblock jupyter-ai.
\newblock \url{https://github.com/jupyterlab/jupyter-ai}.

\bibitem[{Kaplan et~al.(2020)Kaplan, McCandlish, Henighan, Brown, Chess, Child,
  Gray, Radford, Wu, and Amodei}]{kaplan2020scaling}
Jared Kaplan, Sam McCandlish, Tom Henighan, Tom~B Brown, Benjamin Chess, Rewon
  Child, Scott Gray, Alec Radford, Jeffrey Wu, and Dario Amodei. 2020.
\newblock Scaling laws for neural language models.
\newblock \emph{arXiv preprint arXiv:2001.08361}.

\bibitem[{Lai et~al.(2023)Lai, Li, Wang, Zhang, Zhong, Zettlemoyer, Yih, Fried,
  Wang, and Yu}]{lai2023ds}
Yuhang Lai, Chengxi Li, Yiming Wang, Tianyi Zhang, Ruiqi Zhong, Luke
  Zettlemoyer, Wen-tau Yih, Daniel Fried, Sida Wang, and Tao Yu. 2023.
\newblock Ds-1000: A natural and reliable benchmark for data science code
  generation.
\newblock In \emph{International Conference on Machine Learning}, pages
  18319--18345. PMLR.

\bibitem[{Li et~al.(2023{\natexlab{a}})Li, Hammoud, Itani, Khizbullin, and
  Ghanem}]{li2023camel}
Guohao Li, Hasan Abed Al~Kader Hammoud, Hani Itani, Dmitrii Khizbullin, and
  Bernard Ghanem. 2023{\natexlab{a}}.
\newblock Camel: Communicative agents for" mind" exploration of large scale
  language model society.
\newblock \emph{arXiv preprint arXiv:2303.17760}.

\bibitem[{Li et~al.(2023{\natexlab{b}})Li, Li, Li, and Jin}]{li2023structured}
Jia Li, Ge~Li, Yongmin Li, and Zhi Jin. 2023{\natexlab{b}}.
\newblock Structured chain-of-thought prompting for code generation.
\newblock \emph{arXiv preprint arXiv:2305.06599}.

\bibitem[{Liu et~al.(2022)Liu, Shen, Zhang, Dolan, Carin, and
  Chen}]{liu2022makes}
Jiachang Liu, Dinghan Shen, Yizhe Zhang, William~B Dolan, Lawrence Carin, and
  Weizhu Chen. 2022.
\newblock What makes good in-context examples for gpt-3?
\newblock In \emph{Proceedings of Deep Learning Inside Out (DeeLIO 2022): The
  3rd Workshop on Knowledge Extraction and Integration for Deep Learning
  Architectures}, pages 100--114.

\bibitem[{Nori et~al.(2023)Nori, Lee, Zhang, Carignan, Edgar, Fusi, King,
  Larson, Li, Liu, Luo, McKinney, Ness, Poon, Qin, Usuyama, White, and
  Horvitz}]{nori2023can}
Harsha Nori, Yin~Tat Lee, Sheng Zhang, Dean Carignan, Richard Edgar, Nicolo
  Fusi, Nicholas King, Jonathan Larson, Yuanzhi Li, Weishung Liu, Renqian Luo,
  Scott~Mayer McKinney, Robert~Osazuwa Ness, Hoifung Poon, Tao Qin, Naoto
  Usuyama, Chris White, and Eric Horvitz. 2023.
\newblock \href
  {https://www.microsoft.com/en-us/research/publication/can-generalist-foundation-models-outcompete-special-purpose-tuning-case-study-in-medicine/}
  {Can generalist foundation models outcompete special-purpose tuning? case
  study in medicine}.

\bibitem[{OpenAI(2023)}]{openai2023gpt4}
OpenAI. 2023.
\newblock \href {http://arxiv.org/abs/2303.08774} {Gpt-4 technical report}.

\bibitem[{Qian et~al.(2023{\natexlab{a}})Qian, Cong, Liu, Yang, Chen, Su, Dang,
  Li, Xu, Li, Liu, and Sun}]{qian2023communicative}
Chen Qian, Xin Cong, Wei Liu, Cheng Yang, Weize Chen, Yusheng Su, Yufan Dang,
  Jiahao Li, Juyuan Xu, Dahai Li, Zhiyuan Liu, and Maosong Sun.
  2023{\natexlab{a}}.
\newblock \href {http://arxiv.org/abs/2307.07924} {Communicative agents for
  software development}.

\bibitem[{Qian et~al.(2023{\natexlab{b}})Qian, Dang, Li, Liu, Chen, Yang, Liu,
  and Sun}]{qian2023experiential}
Chen Qian, Yufan Dang, Jiahao Li, Wei Liu, Weize Chen, Cheng Yang, Zhiyuan Liu,
  and Maosong Sun. 2023{\natexlab{b}}.
\newblock \href {http://arxiv.org/abs/2312.17025} {Experiential co-learning of
  software-developing agents}.

\bibitem[{Roziere et~al.(2023)Roziere, Gehring, Gloeckle, Sootla, Gat, Tan,
  Adi, Liu, Remez, Rapin et~al.}]{roziere2023code}
Baptiste Roziere, Jonas Gehring, Fabian Gloeckle, Sten Sootla, Itai Gat,
  Xiaoqing~Ellen Tan, Yossi Adi, Jingyu Liu, Tal Remez, J{\'e}r{\'e}my Rapin,
  et~al. 2023.
\newblock Code llama: Open foundation models for code.
\newblock \emph{arXiv preprint arXiv:2308.12950}.

\bibitem[{Shen et~al.(2023)Shen, Song, Tan, Li, Lu, and
  Zhuang}]{shen2023hugginggpt}
Yongliang Shen, Kaitao Song, Xu~Tan, Dongsheng Li, Weiming Lu, and Yueting
  Zhuang. 2023.
\newblock \href {https://openreview.net/forum?id=yHdTscY6Ci} {Hugging{GPT}:
  Solving {AI} tasks with chat{GPT} and its friends in hugging face}.
\newblock In \emph{Thirty-seventh Conference on Neural Information Processing
  Systems}.

\bibitem[{shroominic(2023)}]{codeinterpreterapi}
shroominic. 2023.
\newblock codeinterpreter-api.
\newblock \url{https://github.com/shroominic/codeinterpreter-api}.

\bibitem[{Significant-Gravitas(2023)}]{autogpt}
Significant-Gravitas. 2023.
\newblock Autogpt.
\newblock \url{https://github.com/Significant-Gravitas/AutoGPT}.

\bibitem[{Sui et~al.(2024)Sui, Zhou, Zhou, Han, and Zhang}]{sui2024table}
Yuan Sui, Mengyu Zhou, Mingjie Zhou, Shi Han, and Dongmei Zhang. 2024.
\newblock \href
  {https://www.microsoft.com/en-us/research/publication/table-meets-llm-can-large-language-models-understand-structured-table-data-a-benchmark-and-empirical-study/}
  {Table meets llm: Can large language models understand structured table data?
  a benchmark and empirical study}.
\newblock In \emph{The 17th ACM International Conference on Web Search and Data
  Mining (WSDM '24)}.

\bibitem[{Team et~al.(2023)Team, Anil, Borgeaud, Wu, Alayrac, Yu, Soricut,
  Schalkwyk, Dai, Hauth et~al.}]{team2023gemini}
Gemini Team, Rohan Anil, Sebastian Borgeaud, Yonghui Wu, Jean-Baptiste Alayrac,
  Jiahui Yu, Radu Soricut, Johan Schalkwyk, Andrew~M Dai, Anja Hauth, et~al.
  2023.
\newblock Gemini: a family of highly capable multimodal models.
\newblock \emph{arXiv preprint arXiv:2312.11805}.

\bibitem[{Touvron et~al.(2023)Touvron, Martin, Stone, Albert, Almahairi,
  Babaei, Bashlykov, Batra, Bhargava, Bhosale et~al.}]{touvron2023llama}
Hugo Touvron, Louis Martin, Kevin Stone, Peter Albert, Amjad Almahairi, Yasmine
  Babaei, Nikolay Bashlykov, Soumya Batra, Prajjwal Bhargava, Shruti Bhosale,
  et~al. 2023.
\newblock Llama 2: Open foundation and fine-tuned chat models.
\newblock \emph{arXiv preprint arXiv:2307.09288}.

\bibitem[{Tu et~al.(2023)Tu, Zou, Su, and Zhang}]{tu2023should}
Xinming Tu, James Zou, Weijie~J Su, and Linjun Zhang. 2023.
\newblock What should data science education do with large language models?
\newblock \emph{arXiv preprint arXiv:2307.02792}.

\bibitem[{Wei et~al.(2022)Wei, Wang, Schuurmans, Bosma, Xia, Chi, Le, Zhou
  et~al.}]{wei2022chain}
Jason Wei, Xuezhi Wang, Dale Schuurmans, Maarten Bosma, Fei Xia, Ed~Chi, Quoc~V
  Le, Denny Zhou, et~al. 2022.
\newblock Chain-of-thought prompting elicits reasoning in large language
  models.
\newblock \emph{Advances in Neural Information Processing Systems},
  35:24824--24837.

\bibitem[{Wu et~al.(2023)Wu, Bansal, Zhang, Wu, Zhang, Zhu, Li, Jiang, Zhang,
  and Wang}]{wu2023autogen}
Qingyun Wu, Gagan Bansal, Jieyu Zhang, Yiran Wu, Shaokun Zhang, Erkang Zhu,
  Beibin Li, Li~Jiang, Xiaoyun Zhang, and Chi Wang. 2023.
\newblock Autogen: Enabling next-gen llm applications via multi-agent
  conversation framework.
\newblock \emph{arXiv preprint arXiv:2308.08155}.

\bibitem[{yoheinakajima(2023)}]{babyagi}
yoheinakajima. 2023.
\newblock babyagi.
\newblock \url{https://github.com/yoheinakajima/babyagi}.

\bibitem[{Yu et~al.(2018)Yu, Zhang, Yang, Yasunaga, Wang, Li, Ma, Li, Yao,
  Roman et~al.}]{yu2018spider}
Tao Yu, Rui Zhang, Kai Yang, Michihiro Yasunaga, Dongxu Wang, Zifan Li, James
  Ma, Irene Li, Qingning Yao, Shanelle Roman, et~al. 2018.
\newblock Spider: A large-scale human-labeled dataset for complex and
  cross-domain semantic parsing and text-to-sql task.
\newblock \emph{arXiv preprint arXiv:1809.08887}.

\bibitem[{Zan et~al.(2022)Zan, Chen, Yang, Lin, Kim, Guan, Wang, Chen, and
  Lou}]{zan2022cert}
Daoguang Zan, Bei Chen, Dejian Yang, Zeqi Lin, Minsu Kim, Bei Guan, Yongji
  Wang, Weizhu Chen, and Jian-Guang Lou. 2022.
\newblock {CERT}: Continual pre-training on sketches for library-oriented code
  generation.
\newblock In \emph{The 2022 International Joint Conference on Artificial
  Intelligence}.

\bibitem[{Zhang et~al.(2023{\natexlab{a}})Zhang, Zhang, Ren, Li, and
  Yang}]{zhang2023mlcopilot}
Lei Zhang, Yuge Zhang, Kan Ren, Dongsheng Li, and Yuqing Yang.
  2023{\natexlab{a}}.
\newblock \href {http://arxiv.org/abs/2304.14979} {Mlcopilot: Unleashing the
  power of large language models in solving machine learning tasks}.

\bibitem[{Zhang et~al.(2023{\natexlab{b}})Zhang, Shen, Lu, and
  Zhuang}]{zhang2023data}
Wenqi Zhang, Yongliang Shen, Weiming Lu, and Yueting Zhuang.
  2023{\natexlab{b}}.
\newblock Data-copilot: Bridging billions of data and humans with autonomous
  workflow.
\newblock \emph{arXiv preprint arXiv:2306.07209}.

\end{thebibliography}

\newpage

\onecolumn

\appendix

\section{Validator Implementations}
\label{sec:validator-implementations}

In \autoref{tab:validator-usages}, we list the currently supported validator implementations and the purpose for each of them.
We also show how many times each validator has appeared in the four benchmarks.

\begin{table}[h]
\small
\centering
\begin{tabular}{llp{0.5\textwidth}l}
\toprule
Name         & Alias      & Purpose                                                                               & \# \\
\midrule
Crash        & \texttt{error} & Fail if the generated code crashes.                                                   & 825 \\
Return-Val   & \texttt{execute\_result} & Fail if the executed result of   generated code is not expected.                      & 796 \\
Variables    & \texttt{namespace\_check} & Fail if some variables are not   correctly created   or modified.                     & 276 \\
Unit-test    & \texttt{table\_test} & The defined function fail in at least   one of the test cases.                        & 136 \\
ModelEval    & \texttt{model} & Fail if the defined model does not   satisfy the criteria.                            & 26 \\
Console      & \texttt{stream\_output} & Fail if the console output is not   expected.                                         & 1 \\
AnswerInCode & \texttt{answer\_in\_source} & Succeed if the answer to the query   is shown within the generated code itself.    & 825 \\
Intact       & \texttt{namespace\_intact} & Fail if some variables are unexpectedly   modified, violating intactness constraints. & 825 \\
And          & \texttt{and} & Fail if at least one of the   sub-validators fails.                                    & 825 \\
Or           & \texttt{or} & Succeed if at least one of the   sub-validators succeeds.                              & 825 \\
\bottomrule
\end{tabular}
\caption{Supported validators and their usage counts.}
\label{tab:validator-usages}
\end{table}

\section{Supplementary Evaluations}

\subsection{Introduction to Benchmarked Data Science Agents}
\label{sec:intro-benchmarked-agents}

We briefly introduce the benchmarked data science agents as below.

\begin{itemize}
    \item Chapyter~\cite{chapyter}: A JupyterLab extension translating natural language intentions into Python code with automatic execution. It generates codes based on some predefined examples as well as the conversation history.
    \item ChatDev~\cite{qian2023communicative,qian2023experiential}: A software development framework that operates through the communication between multiple agents, all powered by LLMs. It is non-trivial to adapt ChatDev into an interactive coding agent, thus we only tested it on DSEval-LeetCode.
    \item CoML~\cite{zhang2023mlcopilot}: An interactive coding assistant specifically built for the assistance of data scientists and machine learning practitioners. It has incorporated few-shot examples~\cite{brown2020language}, session variable representations, and code history into the prompt, and also implemented an auto-fixer in case of errors.
    \item Code Interpreter API~\cite{codeinterpreterapi}: An open-sourced implementation of ChatGPT code interpreter. It uses a natural language chatbot as its primary interface. The code executor functions as an external tool.
    \item Jupyter-AI~\cite{jupyterai}: A helpful tool for calling LLMs within a notebook. The generation is purely based on history calls are does not rely on contextual information such as session variables.
\end{itemize}

\subsection{Error Reason Analysis}

\begin{figure*}[h]
\centering
\includegraphics[width=\textwidth]{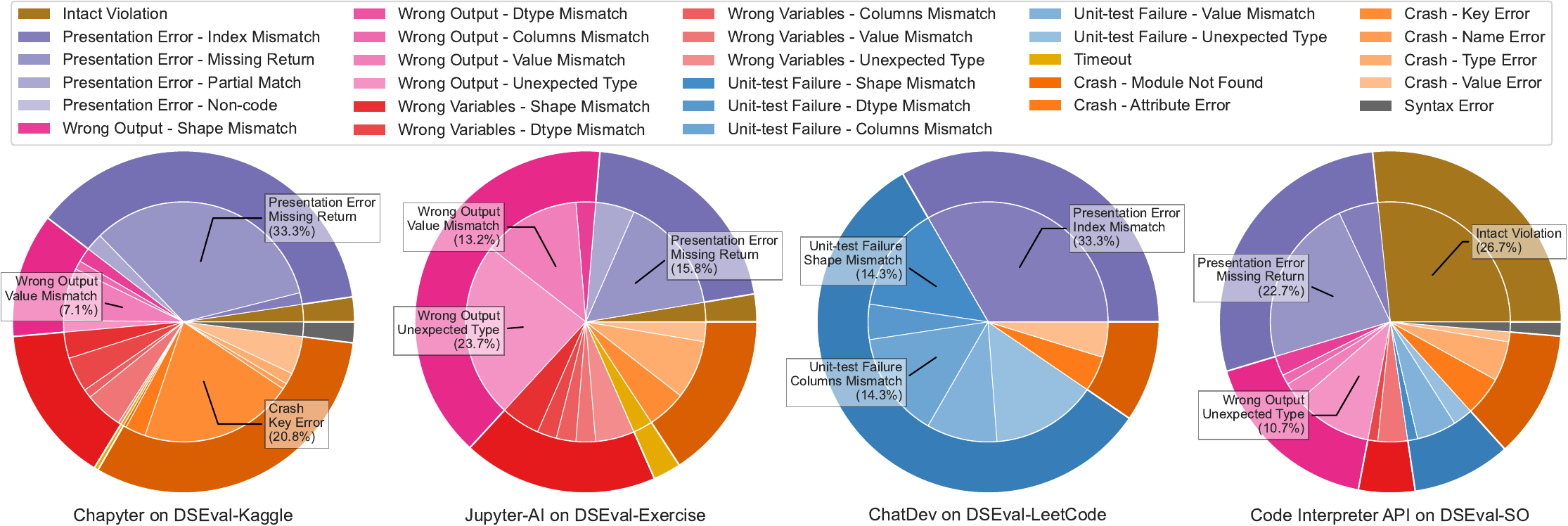}
\caption{A catalog of all error reasons supported by DSEval (full explanations in \autoref{sec:appendix-verdict-catalog}). The error causes of a selected subset of data science agents on the benchmarks are visualized in the pie charts.}
\label{fig:verdict-pie-chart-full}
\end{figure*}

From \autoref{fig:verdict-pie-chart-full}, we can see that although Chapyter on DSEval-Kaggle and ChatDev on DSEval-Kaggle both suffer from presentation error, one is primarily due to missing return (e.g., using ``\texttt{print()}'' instead of ``\texttt{return}'' to show the output), the other is due to index match (e.g., naming the columns with a wrong name).
The error cause of Jupyter-AI is rather diverse, with ``wrong output'' being the dominant cause.
Code Interpreter API on DSEval-SO often triggers intact violation, as the framework has a tendency to perform inplace modifications to existing variables.

A detailed explanation of each error reason can be found in \autoref{sec:appendix-verdict-catalog}.

\subsection{Prompt Techniques}

We incorporate various prompt techniques that are commonly used for different tasks into CoML for evaluation.
Our goal is to identify the strengths and weaknesses of these techniques under data science scenarios.

\paragraph{Chain-of-thought.}
CoT~\cite{wei2022chain} is a popular method for generating prompts that can handle various tasks.
However, as shown in \autoref{tab:compare-cot}, CoT does not perform well on DSEval benchmarks as expected.
A possible explanation is that the code itself already has a logical structure and does not require additional chain-of-thoughts.
This result is consistent with recent works such as SCoT~\cite{li2023structured}, which introduces CoT variants for code generation tasks.
However, since most data science code lacks the ``structure'' of loops and conditions, adapting the method is challenging and we leave it as future work.

\begin{table}[h]
\centering
\small
\begin{tabular}{ccccc}
\toprule
Prompt & Kaggle & Exercise & LeetCode & SO \\
\midrule
CoML & \bfseries 59.8 & 78.6 & 42.5 & \bfseries 78.2 \\
CoML + COT & 57.8 & \bfseries 80.2 & \bfseries 45.0 & 76.2 \\
\bottomrule
\end{tabular}
\caption{Comparison of CoML w/ and w/o CoT.}
\label{tab:compare-cot}
\end{table}

\paragraph{Few-shot prompting.}
Few-shot prompting~\cite{kaplan2020scaling} is a method that uses demonstrations in prompts to help the model learn from the context.
CoML uses a 5-shot prompt (5 demonstrations) by default to improve the quality of its generation.
Few-shot prompting has the drawback of using more tokens (around 1k for 5 demonstrations).
We want to see what happens when we use less demonstrations in the prompt.

\begin{figure}[h]
\centering
\begin{subfigure}[b]{0.315\textwidth}
\includegraphics[width=\linewidth]{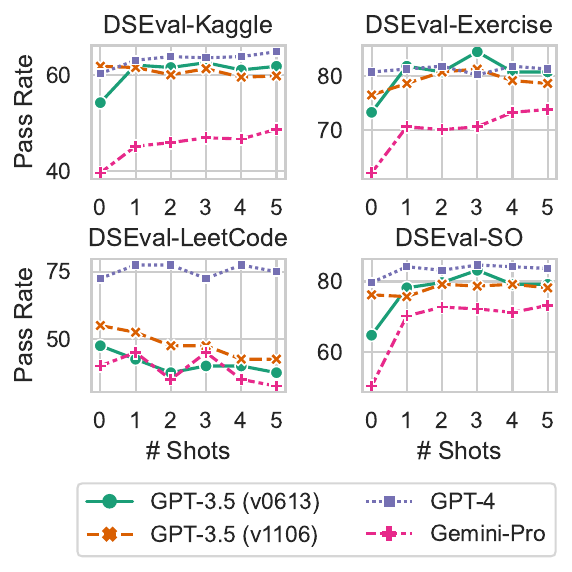}
\caption{With different LLMs.}
\label{fig:compare-shots-model}
\end{subfigure}
\hfill
\begin{subfigure}[b]{0.33\textwidth}
\includegraphics[width=\linewidth]{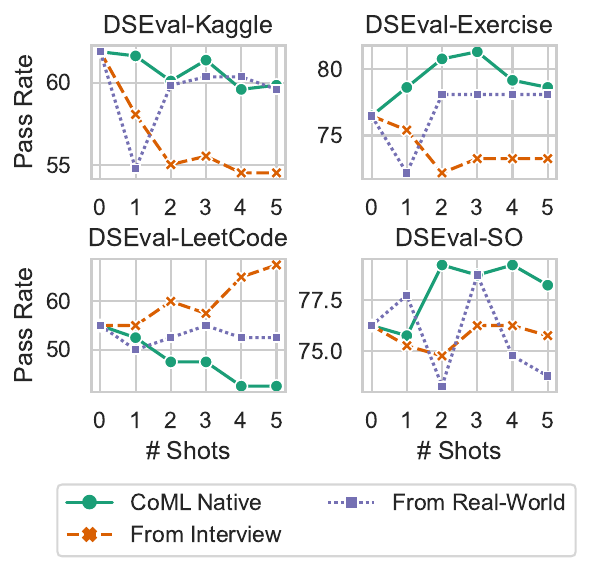}
\caption{Different sets of demonstrations.}
\label{fig:compare-shots-prompt}
\end{subfigure}
\hfill
\begin{subfigure}[b]{0.315\textwidth}
\includegraphics[width=\linewidth]{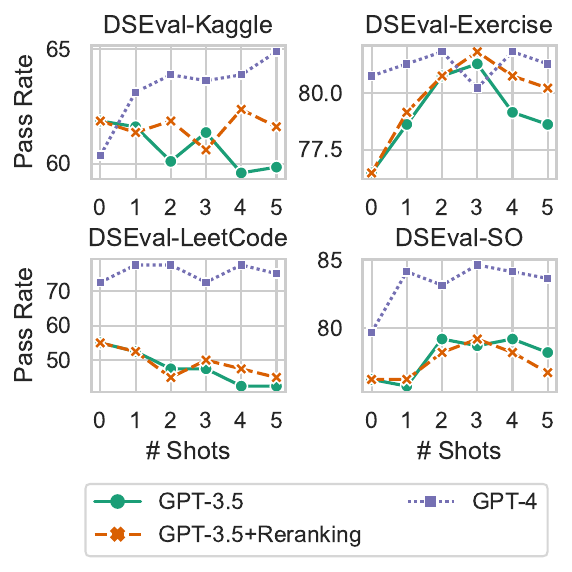}
\caption{Rerank the examples by relevance.}
\label{fig:compare-shots-retrieval}
\end{subfigure}
\caption{Impact of number of examples used for few-shot prompting.}
\label{fig:compare-shots}
\end{figure}

In \autoref{fig:compare-shots-model}, we use a simple strategy, that is to keep the first $k$ demonstrations in the order of appearance, where $k$ is the number of demonstrations to keep.
We ran the experiment with different backend LLMs, including two versions of GPT-3.5, GPT-4, and Gemini-Pro.
The results show that the performance tends to get better with more shots (i.e., demonstrations).
But there are also some exceptions.
For instance, the pass rate of GPT-3.5 keeps going down on DSEval-LeetCode.
On DSEval-Kaggle, the pass rate also fluctuates and the zero-shot performance is not worse than more shots.

We hypothesize that this phenomenon is because of a misalignment between the demonstrations and the benchmarks.
The demonstrations in CoML are made with toy datasets and problems, which might not match the problems in each benchmark.
In \autoref{fig:compare-shots-prompt}, we manually created two more sets of demonstrations.
One is from real-world situations such as data processing and model training.
The other is from interview questions, from platforms like LeetCode.
We made sure that the demonstrations did not overlap with any problem in the benchmarks.
As shown in \autoref{fig:compare-shots-prompt}, with demonstrations from interviews, DSEval-LeetCode benefits a lot from demonstrations.
However, this set of demonstrations does not work well for other benchmarks.
Demonstrations from real-world have an unstable performance and generally not satisfactory, implying that choosing the right demonstrations is a challenging issue in this scenario.

\paragraph{Reranking.}
Another well-known technique that is often used in conjunction with few-shot prompting is reordering the demonstrations~\cite{liu2022makes,nori2023can}, which is also known as ``similarity-based example selector'' or ``$k$NN-curated dynamic few-shot exemplar''.
In our setting, we reorder the demonstrations by their cosine similarity (on ``text-embedding-ada-002'') with the incoming query at inference time, and select the top $k$ demonstrations as the $k$ shots.
The outcomes are presented in \autoref{fig:compare-shots-retrieval}.
Reordering the demonstrations is generally helpful for the performance, but the improvement is marginal.
The performance is still much lower than a more powerful LLM (e.g., GPT-4).

\subsection{Comparison of Different LLMs}
\label{sec:compare-models-more}

As can be seen in \autoref{tab:compare-models}, less capable models tend to suffer more from stricter evaluation settings (e.g., with error propagation).
More capable models are also better at following instructions to preserve intactness or follow the desired format.
For CodeLlama and Gemini, the pass rates can improve up to 7$\sim$9\% when loosing the presentation error, but for GPT-3.5 and GPT-4 the improvement is much smaller.

\begin{table*}[h]
\centering
\resizebox{\textwidth}{!}{\begin{tabular}{c|cccc|cccc|ccc|ccc}
\toprule
\multirow{2}*{Model} & \multicolumn{4}{c|}{DSEval-Kaggle} & \multicolumn{4}{c|}{DSEval-Exercise} & \multicolumn{3}{c|}{DSEval-LeetCode} & \multicolumn{3}{c}{DSEval-SO} \\
& \begin{tabular}{@{}c@{}}Pass \\ Rate\end{tabular} & \begin{tabular}{@{}c@{}}Error \\ Prop\end{tabular} & \begin{tabular}{@{}c@{}}w/o \\ Intact\end{tabular} & \begin{tabular}{@{}c@{}}w/o \\ PE\end{tabular}  & \begin{tabular}{@{}c@{}}Pass \\ Rate\end{tabular} & \begin{tabular}{@{}c@{}}Error \\ Prop\end{tabular} & \begin{tabular}{@{}c@{}}w/o \\ Intact\end{tabular} & \begin{tabular}{@{}c@{}}w/o \\ PE\end{tabular}  & \begin{tabular}{@{}c@{}}Pass \\ Rate\end{tabular} & \begin{tabular}{@{}c@{}}w/o \\ Intact\end{tabular} & \begin{tabular}{@{}c@{}}w/o \\ PE\end{tabular}  & \begin{tabular}{@{}c@{}}Pass \\ Rate\end{tabular} & \begin{tabular}{@{}c@{}}w/o \\ Intact\end{tabular} & \begin{tabular}{@{}c@{}}w/o \\ PE\end{tabular} \\
\midrule
CodeLlama-7B & {\cellcolor[HTML]{DED2E0}} \color[HTML]{000000} 30.6 & {\cellcolor[HTML]{E7E1E9}} \color[HTML]{000000} 24.5 & {\cellcolor[HTML]{DCD0DF}} \color[HTML]{000000} 31.6 & {\cellcolor[HTML]{D3C1D7}} \color[HTML]{000000} 37.9  & {\cellcolor[HTML]{DED2E0}} \color[HTML]{000000} 52.9 & {\cellcolor[HTML]{EFEDEF}} \color[HTML]{000000} 44.4 & {\cellcolor[HTML]{DDD0DF}} \color[HTML]{000000} 53.5 & {\cellcolor[HTML]{D6C6DA}} \color[HTML]{000000} 56.7  & {\cellcolor[HTML]{F1F0F1}} \color[HTML]{000000} 12.5 & {\cellcolor[HTML]{F1F0F1}} \color[HTML]{000000} 12.5 & {\cellcolor[HTML]{E6DEE7}} \color[HTML]{000000} 22.5  & {\cellcolor[HTML]{F1F0F1}} \color[HTML]{000000} 45.5 & {\cellcolor[HTML]{EEECEF}} \color[HTML]{000000} 47.0 & {\cellcolor[HTML]{E3DAE5}} \color[HTML]{000000} 53.0 \\
CodeLlama-34B & {\cellcolor[HTML]{E2D9E4}} \color[HTML]{000000} 27.8 & {\cellcolor[HTML]{F1F0F1}} \color[HTML]{000000} 17.9 & {\cellcolor[HTML]{E1D7E3}} \color[HTML]{000000} 28.8 & {\cellcolor[HTML]{D0BDD4}} \color[HTML]{000000} 39.4  & {\cellcolor[HTML]{E3DBE5}} \color[HTML]{000000} 50.3 & {\cellcolor[HTML]{F1F0F1}} \color[HTML]{000000} 43.3 & {\cellcolor[HTML]{E3DBE5}} \color[HTML]{000000} 50.3 & {\cellcolor[HTML]{D0BCD4}} \color[HTML]{000000} 59.9  & {\cellcolor[HTML]{ECE8EC}} \color[HTML]{000000} 17.5 & {\cellcolor[HTML]{ECE8EC}} \color[HTML]{000000} 17.5 & {\cellcolor[HTML]{E3DAE5}} \color[HTML]{000000} 25.0  & {\cellcolor[HTML]{ECE9ED}} \color[HTML]{000000} 48.0 & {\cellcolor[HTML]{ECE8EC}} \color[HTML]{000000} 48.5 & {\cellcolor[HTML]{DDD1E0}} \color[HTML]{000000} 55.9 \\
Gemini-Pro & {\cellcolor[HTML]{C2A7C8}} \color[HTML]{000000} 48.7 & {\cellcolor[HTML]{CCB7D1}} \color[HTML]{000000} 41.9 & {\cellcolor[HTML]{C2A6C8}} \color[HTML]{000000} 49.0 & {\cellcolor[HTML]{B795BE}} \color[HTML]{F1F1F1} 56.1  & {\cellcolor[HTML]{B390BB}} \color[HTML]{F1F1F1} 73.8 & {\cellcolor[HTML]{C0A4C7}} \color[HTML]{000000} 67.4 & {\cellcolor[HTML]{B390BB}} \color[HTML]{F1F1F1} 73.8 & {\cellcolor[HTML]{AD86B6}} \color[HTML]{F1F1F1} 77.0  & {\cellcolor[HTML]{DACCDD}} \color[HTML]{000000} 32.5 & {\cellcolor[HTML]{DACCDD}} \color[HTML]{000000} 32.5 & {\cellcolor[HTML]{CBB5D0}} \color[HTML]{000000} 45.0  & {\cellcolor[HTML]{BB9CC2}} \color[HTML]{F1F1F1} 73.3 & {\cellcolor[HTML]{BB9CC2}} \color[HTML]{F1F1F1} 73.3 & {\cellcolor[HTML]{AF8AB8}} \color[HTML]{F1F1F1} 79.2 \\
GPT-3.5 & {\cellcolor[HTML]{B18CB9}} \color[HTML]{F1F1F1} 59.8 & {\cellcolor[HTML]{B693BD}} \color[HTML]{F1F1F1} 56.8 & {\cellcolor[HTML]{AF89B7}} \color[HTML]{F1F1F1} 61.1 & {\cellcolor[HTML]{AB83B4}} \color[HTML]{F1F1F1} 63.6  & {\cellcolor[HTML]{A981B3}} \color[HTML]{F1F1F1} 78.6 & {\cellcolor[HTML]{A981B3}} \color[HTML]{F1F1F1} 78.6 & {\cellcolor[HTML]{A97FB2}} \color[HTML]{F1F1F1} 79.1 & {\cellcolor[HTML]{A478AE}} \color[HTML]{F1F1F1} 81.3  & {\cellcolor[HTML]{CEBAD3}} \color[HTML]{000000} 42.5 & {\cellcolor[HTML]{CEBAD3}} \color[HTML]{000000} 42.5 & {\cellcolor[HTML]{B695BE}} \color[HTML]{F1F1F1} 62.5  & {\cellcolor[HTML]{B18DB9}} \color[HTML]{F1F1F1} 78.2 & {\cellcolor[HTML]{AE88B7}} \color[HTML]{F1F1F1} 79.7 & {\cellcolor[HTML]{AE88B7}} \color[HTML]{F1F1F1} 79.7 \\
GPT-4 & {\cellcolor[HTML]{A980B2}} \color[HTML]{F1F1F1} 64.9 & {\cellcolor[HTML]{B28EBA}} \color[HTML]{F1F1F1} 59.3 & {\cellcolor[HTML]{A57AAF}} \color[HTML]{F1F1F1} 67.4 & {\cellcolor[HTML]{A275AC}} \color[HTML]{F1F1F1} 69.7  & {\cellcolor[HTML]{A478AE}} \color[HTML]{F1F1F1} 81.3 & {\cellcolor[HTML]{AB83B4}} \color[HTML]{F1F1F1} 78.1 & {\cellcolor[HTML]{A478AE}} \color[HTML]{F1F1F1} 81.3 & {\cellcolor[HTML]{A275AC}} \color[HTML]{F1F1F1} 82.4  & {\cellcolor[HTML]{A87EB1}} \color[HTML]{F1F1F1} 75.0 & {\cellcolor[HTML]{A87EB1}} \color[HTML]{F1F1F1} 75.0 & {\cellcolor[HTML]{A275AC}} \color[HTML]{F1F1F1} 80.0  & {\cellcolor[HTML]{A77CB0}} \color[HTML]{F1F1F1} 83.7 & {\cellcolor[HTML]{A579AE}} \color[HTML]{F1F1F1} 84.7 & {\cellcolor[HTML]{A275AC}} \color[HTML]{F1F1F1} 86.1 \\
\bottomrule
\end{tabular}%
}
\caption{Comparison of different LLMs. The metrics are: pass rate, pass rate with error propagation, pass rate without the constraint of intact violation, and pass rate without considering presentation error.}
\label{tab:compare-models}
\end{table*}

\section{Verdict Catalog}
\label{sec:appendix-verdict-catalog}

The following table summarize all the verdicts and subverdicts supported in DSEval.
We refer to the code generated by the benchmarked agent as ``\emph{submission}'' and the oracle code as ``\emph{reference}''.

\begin{small}
\begin{longtable}{|l|l|p{0.25\textwidth}|p{0.3\textwidth}|}
\hline
Verdict & Sub-verdict & Explanation & Example \\ \hline
\multicolumn{2}{|l|}{Correct} & Correct. & \\ \hline
\multicolumn{2}{|l|}{Intact Violation} & The submission violates the constraints of not modifying, updating or deleting existing variables unless necessary. & Q: What is the most dangerous decade to live in the US? Write it in the format of ``19XXs'' or ``20XXs''.

\begin{minipage}[t]{\linewidth}
\begin{minted}[breaklines,fontsize=\scriptsize]{python}
crimes['Total_Crimes'] = crimes.iloc[:, 1:].sum(axis=1)
most_dangerous_decade = crimes['Total_Crimes'].idxmax()
most_dangerous_decade.strftime("%Ys")
\end{minted}
\end{minipage}
\\ \hline
\multirow{4}*{Presentation Error} & Index Mismatch & Only for DataFrame / Series outputs. The submission DataFrame / Series is correct, but has the wrong column names, incorrect index, or not properly sorted. & Count the number of fatalities for each year. Return a Series with ``Year'' as the index and ``Number of Fatalities'' as the values.

\begin{minipage}[t]{\linewidth}
\begin{minted}[breaklines,fontsize=\scriptsize]{python}
fatalities['date_of_event'] \
  .dt.year.value_counts() \
  .rename("Number of Fatalities")
\end{minted}
\end{minipage}
\\ \cline{2-4}
 & Missing Return & The submission output is printed to the console output rather than put into the desired returning value. & Q: Show the first rows of the dataset.

\begin{minipage}[t]{\linewidth}
\begin{minted}[breaklines,fontsize=\scriptsize]{python}
print(df.head())
\end{minted}
\end{minipage}
\\ \cline{2-4}
 & Partial Match & The desired output can be partially found within the submission output. For example, the reference output is a subset DataFrame of the submission, or the index of the submitted series, etc. & Q: List the names of the top 10 industries that have produced the most billionaires.

\begin{minipage}[t]{\linewidth}
\begin{minted}[breaklines,fontsize=\scriptsize]{python}
billionaires.groupby('industries') \
  ['personName'].count() \
  .sort_values(ascending=False). \
  head(10)
\end{minted}
\end{minipage}
\\ \cline{2-4}
 & Non-code & The submission generates plain texts (rather than code) to answer the query. & Q: What is the number of columns in the dataset?

\begin{minipage}[t]{\linewidth}
\begin{minted}[breaklines,fontsize=\scriptsize]{python}
The number of columns in the dataset is 5.
\end{minted}
\end{minipage}
\\ \hline
\multirow{6}*{Wrong Output} & Shape Mismatch & Output is wrong in the shape of the DataFrame or array. & Q: Select all columns except the last 3.

\begin{minipage}[t]{\linewidth}
\begin{minted}[breaklines,fontsize=\scriptsize]{python}
euro12.iloc[: , 0:7]
\end{minted}
\end{minipage}
\\ \cline{2-4}
 & Dtype Mismatch & Submission output is a DataFrame / Series and is wrong in the data type. & Q: Encode the categorical feature with label encoder and transform it into float.

\begin{minipage}[t]{\linewidth}
\begin{minted}[breaklines,fontsize=\scriptsize]{python}
LabelEncoder().fit_transform(x)
\end{minted}
\end{minipage}
\\ \cline{2-4}
 & Columns Mismatch & Submission output is a DataFrame / Series and the column names are not expected. & Q: Remove excessive spaces from the column names. Save the cleaned dataset in-place.
 
\begin{minipage}[t]{\linewidth}
\begin{minted}[breaklines,fontsize=\scriptsize]{python}
netflix.columns = netflix.columns.str.strip()
\end{minted}
\end{minipage}
\\ \cline{2-4}
 & Value Mismatch & Output is wrong in the data itself. & Q: Is there any duplicate dates?

\begin{minipage}[t]{\linewidth}
\begin{minted}[breaklines,fontsize=\scriptsize]{python}
apple.index.duplicated().any()
\end{minted}
\end{minipage}
\\ \cline{2-4}
 & Unexpected Type & & Q: Get a summary with the mean, min, max, std and quartiles of the dataset.

\begin{minipage}[t]{\linewidth}
\begin{minted}[breaklines,fontsize=\scriptsize]{python}
baby_names['Count'].describe()
\end{minted}
\end{minipage}
\\ \cline{2-4}
 & Others & Uncategorized wrong output. & \\ \hline
Wrong Variables & \multicolumn{1}{@{}l|}{\begin{tabular}{l} Shape Mismatch \\ \hline Dtype Mismatch \\ \hline Columns Mismatch \\ \hline Value Mismatch \\ \hline Unexpected Type \\ \hline Others \end{tabular}} &
Variables are incorrect after execution of the submission code. Sub-verdicts are the same as ``Wrong Output''. & 
Q: Add another column called place. The values of place are as follows: Bulbasaur is in park, Caterpie is in forest, Squirtle is in lake, Charmander is in street.

\begin{minipage}[t]{\linewidth}
\begin{minted}[breaklines,fontsize=\scriptsize]{python}
pokemon_col["place"] = [
  "park", "forest", "lake", "street"]
\end{minted}
\end{minipage}
\\ \hline
Unit-test Failure & \multicolumn{1}{@{}l|}{\begin{tabular}{l} Shape Mismatch \\ \hline Dtype Mismatch \\ \hline Columns Mismatch \\ \hline Value Mismatch \\ \hline Unexpected Type \\ \hline Others \end{tabular}} &
The function in the submission did not pass the pre-defined unit-tests. Sub-verdicts are the same as ``Wrong Output''. & 
Q: Write a sentiment prediction function called \texttt{predict\_sentiment}. The function should take a review as input and return the predicted sentiment (``Positive'', ``Negative'', or ``Neutral'') as output.

\begin{minipage}[t]{\linewidth}
\begin{minted}[breaklines,fontsize=\scriptsize]{python}
def predict_sentiment(review):
  words = word_tokenize(review.lower())
  words = [word for word in words if word.isalpha() and word not in stopwords.words('english')]
  features = vectorizer.transform(words)
  return model.predict(features)
\end{minted}
\end{minipage}
\\ \hline
\multicolumn{2}{|l|}{Timeout} & The code fails to finish in the limited time. It could be due to endless loops or inefficiency. & Q: Use grid search to tune the hyperparameters of the random forest classifier. The time limit is 30 seconds.

\begin{minipage}[t]{\linewidth}
\begin{minted}[breaklines,fontsize=\scriptsize]{python}
param_grid = {
  'n_estimators': [100, 200, 300],
  'max_depth': [None, 10, 20],
  'min_samples_split': [2, 5, 10],
  'min_samples_leaf': [1, 5, 10],
  'bootstrap': [True, False],
  'criterion': ['gini', 'entropy']
}

grid_search = GridSearchCV(model, param_grid, cv=5, n_jobs=-1, verbose=1)
grid_search.fit(X_train, y_train)
\end{minted}
\end{minipage}
\\ \hline
\multirow{7}*{Crash} & Module Not Found & Usually when the code fails to import a library. & Q: Conduct a chi-squared test to examine the relationship between ``Gender'' and ``Discount Applied''. Show the chi-squared statistic.

\begin{minipage}[t]{\linewidth}
\begin{minted}[breaklines,fontsize=\scriptsize]{python}
import pandas as pd
import numpy as np
from stats import chi2_contingency
pd.crosstab(shopping['Gender'], shopping['Discount Applied'])
# No module named 'stats'
\end{minted}
\end{minipage}
\\ \cline{2-4}
 & Attribute Error & Usually happens when referencing a non-existing method or attribute. & Q: Fill the missing values with NaN

\begin{minipage}[t]{\linewidth}
\begin{minted}[breaklines,fontsize=\scriptsize]{python}
salaries_growth_rate = \
  salaries_growth_rate \
  .fillna(value=pd.np.nan)
\end{minted}
\end{minipage}
\\ \cline{2-4}
 & Key Error & Usually happens when referencing a non-existing column. & Q: Select the third cell in the row named Arizona

\begin{minipage}[t]{\linewidth}
\begin{minted}[breaklines,fontsize=\scriptsize]{python}
army.loc["Arizona", 2]
\end{minted}
\end{minipage}
\\ \cline{2-4}
 & Name Error & Referencing an undefined variable or using an unimported API. & Q: Calculate the pearson correlation between the final worth and age of billionaires.

\begin{minipage}[t]{\linewidth}
\begin{minted}[breaklines,fontsize=\scriptsize]{python}
df['finalWorth'].corr(df['age'])
# name 'df' is not undefined
\end{minted}
\end{minipage}
\\ \cline{2-4}
 & Type Error & Happens when a type is misused. For example, running numeric operations on string types. & Q: How many products have a unit cost more than \$10.00?

\begin{minipage}[t]{\linewidth}
\begin{minted}[breaklines,fontsize=\scriptsize]{python}
chipo['item_price'] > 10
# '>' not supported between instances of 'str' and 'int'
\end{minted}
\end{minipage}
\\ \cline{2-4}
 & Value Error & Happens when operations can not process certain values. & Q: Compute the correlation of heart attack risk against other numeric features. Sort the factors by the absolute values of the correlation coefficients in descending order.

\begin{minipage}[t]{\linewidth}
\begin{minted}[breaklines,fontsize=\scriptsize]{python}
corr_matrix = heart.corr()
corr_matrix.abs().sort_values(
  ascending=False)
# could not convert string to float: 'BMW7812'
\end{minted}
\end{minipage}
\\ \cline{2-4}
 & Others & Uncategorized Crash. & \\ \hline
\multicolumn{2}{|l|}{Syntax Error} & Code has syntax error. & \\ \hline

\caption{Catalog of verdicts supported by DSEval. In case a solution is problematic from multiple perspectives, the verdicts from the bottom of the table have higher priorities to appear.}
\label{tab:verdict-catalog}
\end{longtable}
\end{small}

\section{Result Visualizer}
\label{sec:result-visualizer}

\begin{figure*}[h]
\includegraphics[width=\textwidth]{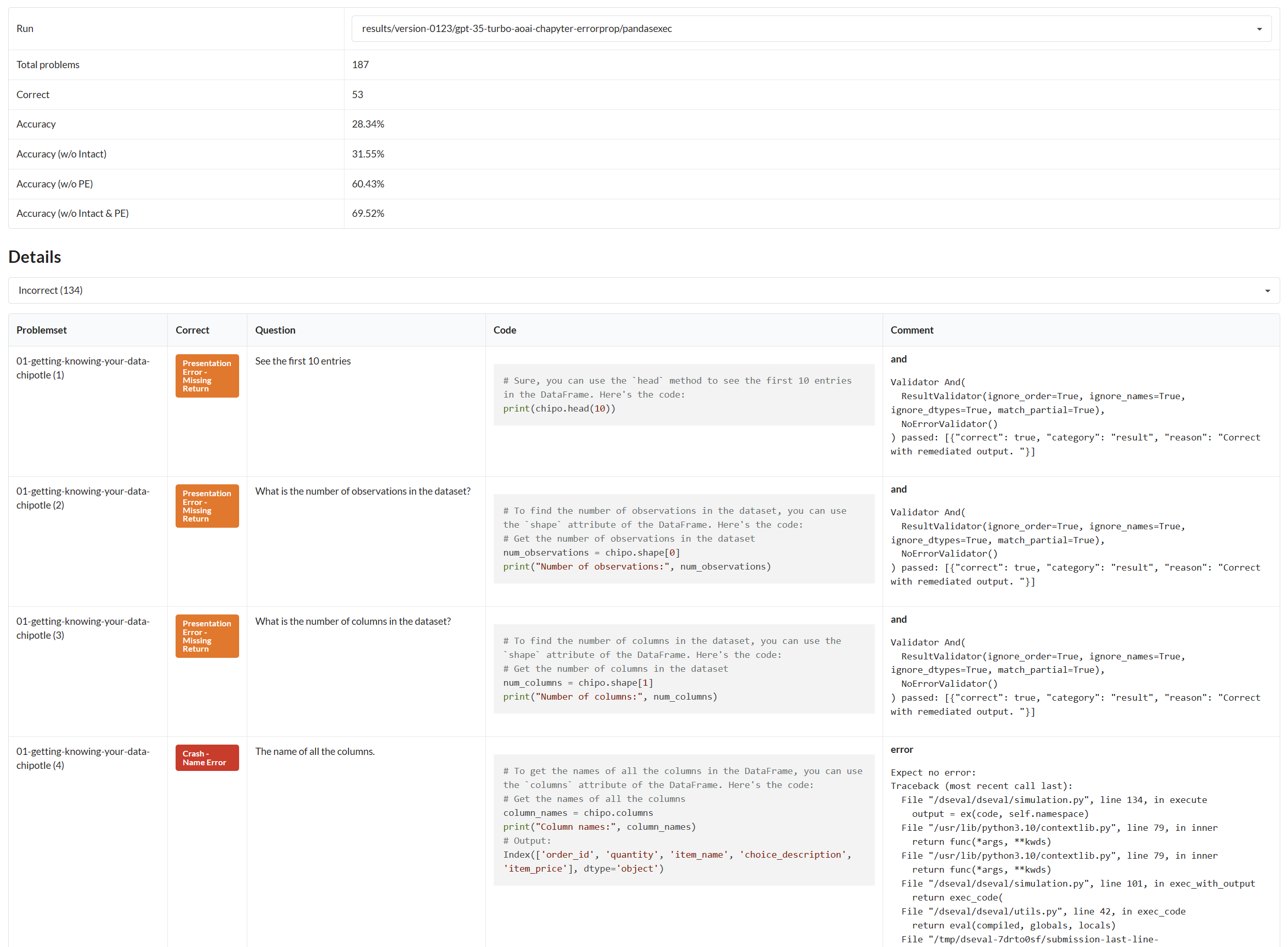}
\caption{Result Visualizer.}
\label{fig:result-visualizer}
\end{figure*}

We build a visualizer accompanying DSEval, to facilitate the examination and diagnosis of the results. A demonstration is shown in \autoref{fig:result-visualizer}.

\section{Benchmark Annotation Details}
\label{sec:benchmark-annotation-details}

\subsection{DSEval-Exercise}

Pandas-exercises~\cite{pandasexercises} contains 252 tutorial-purposed pandas questions organized in 27 notebooks. Each notebook contains multiple mutually correlated questions, featuring one specific theme, which could be filtering, sorting, time series, visualization and etc.
The notebooks can be converted into DSEAL through a simple rule-based conversion script.
We manually clarified some vague questions and corrected the validator configurations for each question to allow for proper error tolerances for some problems.
Visualization problems are discarded due to validators of charts are not implemented, which we left as future work.
We end up selecting 187 problems from 21 problem sets, which we call ``DSEval-Exercise''.

\subsection{DSEval-SO}

CERT~\cite{zan2022cert} presents PandasEval and NumpyEval, which altogether contain 202 pandas and numpy problems, collected from StackOverflow (SO).
The queries are mostly related to usages of pandas or numpy APIs and tricks.
Most answers are very short in terms of the number of operations involved.
The original benchmarks are in the form of code completion.
We manually clarified ambiguous queries and converted them into DSEAL with the help of GPT-4~\cite{openai2023gpt4}.

\subsection{DSEval-LeetCode}

LeetCode has published several dozens of problems targetting data science areas\footnote{\url{https://leetcode.com/problemset/pandas/}}.
By August 2023, 40 of them were available to free-tier accounts, which we crawled and converted into DSEAL with GPT-4.

We prompt GPT-4 to start the converted query by ``Write a function \texttt{\`{}def ...:\`{}}'', followed by the explanation of its inputs and outputs.
We also instruct GPT-4 to include both the problem statement and sample inputs/outputs in our query part as we found that the query could often be ambiguous without the samples.
``\texttt{table\_test}'' validators are used to validate the agents' output as the task of the agents is to write a function that can handle specific kinds of inputs.
Since test cases and standard solutions are not obtainable directly from LeetCode, those parts are also written by GPT-4.
The test cases are meant to cover the samples, scaled inputs, and some corner cases.
To ensure the generated test cases were legal, we also asked the GPT-4 to generate a function to prevent illegal cases from coming into the benchmark.

After the conversion, we first submitted the generated solutions to the LeetCode platform for verification.
38 solutions were successfully submitted (as the other 2 turned out to require premium access), out of which 27 were correct.
We manually fixed 10 out of 11 wrong solutions, while we found the other one mistakenly judged to be incorrect due to an invalid test case on LeetCode.
We then tested the solutions on the auto-generated test cases, and manually corrected the tests, validators, or solutions where the results turned out to be incorrect.
Finally, we used a data science agent (GPT-3.5 with CoML) for a trial run on this benchmark, and cross-checked our verdicts with the verdict on LeetCode online.
We strengthened several test cases where our system returned correct and LeetCode returned otherwise.
We show an example problem in DSEval-LeetCode in \autoref{sec:problem-examples}.

\subsection{DSEval-Kaggle}
\label{sec:dseval-kaggle-prompts}

Here, we detail the annotation costs associated with the DSEval-Kaggle problemsets, building upon the process outlined in the main text.

The dataset comprises 31 problemsets.
During the initial iteration, all 31 were generated without an exemplary problemset for reference.
Human annotators then reviewed and selected one for revision, leaving the remaining 30 for further generation.
This iterative process continued, with two problemsets revised at the second stage, two problemsets revised at the third stage, four problemsets revised at the third stage, until all 31 had undergone revision at the fifth stage.
When the number of few-shot examples exceeded five, random selection was employed.

The total prompt token consumption for sketch generation amounted to 877 thousand tokens, while generating the full problemset consumed 1.44 million tokens.

Initially, human annotators reported requiring approximately one hour to revise each problemset sketch. This involved understanding the dataset content, adjusting interdependencies within the problemset, and occasionally conducting web searches for clarification.
An additional hour was needed to revise the full problemset, including validator configurations and reference code fixes.
Notably, the annotation time decreased significantly in later iterations of the outer loop, dropping to ``20 to 30 minutes'' per problemset.

The final revised ``DSEAL guide'' are shown below.

\begin{tcolorbox}[title=Problem Sketch Instruction]
\small
Your task is to help a teacher design a problemset for an examination. The problemset is to test the students' ability to write Python code to solve data science problems (using numpy and pandas). The dataset that will be used for the problemset is pre-determined and shall be given by the user. You will also be provided a reference that might give you some ideas on what can be done with this dataset, but do not rely on it or copy it. You should write a sketch of the new problemset using the provided dataset. The sketch should include the following information:

\begin{itemize}
    \item The knowledge points of the problemset: what knowledge points or programming skills are tested in the problemset?
    \item A sketch of the problemset: How many subproblems roughly are there in the problemset? what is each subproblem in the problemset about? How are the subproblems related to each other?
\end{itemize}

<DATASET DESCRIPTION>

For the new dataset mentioned above, please design a new problemset that is more difficult and more challenging than all the problemsets above, and write its desired knowledge points and sketch. Please follow the instructions below:

\begin{itemize}
    \item The new problemset should also be different from the existing problemsets, i.e., it should not be a combination of existing problemsets.
    \item The new problemset should cover some new knowledge points or programming skills that are not covered by the existing problemsets.
    \item The problemset should contain roughly 10 - 15 problems.
    \item But try to follow the format of the existing problemsets.
    \item Problems with more logical thinking and reasoning challenges are preferred.
    \item Do not include visualization problems, system design problems, model training problems or open questions as I won't be able to automatically evaluate their results.
    \item Please do not be constrained by the ideas from existing problemsets. You can design a problemset that is novel, creative and interesting.
\end{itemize}
\end{tcolorbox}

\begin{tcolorbox}[title=Full Problemset Writing Instruction]
\small
Your task is to help a teacher design a problemset for an examination. The problemset is to test the students' ability to write Python code to solve data science problems (using numpy and pandas). In particular, you are to write a full problemset based on a scratch.

The desired format is a Python file with multiple cells separated with ``\texttt{\# \%\%}''. The first cell is some preparation code (e.g., import libraries like pandas), and the rest are the tasks. Each task consists of a docstring (containing question and validator) and a code block (containing the reference solution). The docstring is written in YAML, and the code block is written in Python.

Some extra instructions:

\begin{itemize}
\item Data files used in the problems are located under \texttt{inputs/}. You can use them in your problemset.
\item If the sketch contains problems that are ambiguous or do not make sense, you can refine them. You can also add more problems to the problemset.
\item When the sketch gives problem examples like ``such as'', ``e.g.'', ``for example'', etc., you can think of your own problem based on the given data. You don't need to follow the exact concrete problems given in the sketch.
\item When using external data, you should use your knowledge to find the right URL on the Internet. You should write a separate question to read the data from online, and then use the data in the following questions.
\item The result of each subproblem's reference code should ideally be a single value (e.g., a number, a string, a list, a dictionary, a dataframe, etc.). When students submit their code, the result of their code will be compared with the result of the reference code. If the results are the same, the student's code is considered correct. Otherwise, the student's code is considered incorrect.
\item When manipulating the data and creating the features, try to adhere to the style and content of original data. For example, if the data columns are named in camel case, you should also name new columns in camel case. If the data only contains values between 0 and 1, you should not create a new feature that categorizes the data into 0-10, 10-20, etc.
\item To make the comparison above possible, the result of the reference code should be the one and only possible answer to the question. Therefore, the question should be specific enough to have only one possible answer. For example, instead of asking ``Provide a summary of the dataset'', you should ask ``What is the mean, std of the temperature anomalies of \texttt{dataset\_a}? Put them in a tuple'', or ``return the results in a dataframe with columns \texttt{mean} and \texttt{std}'', or ``show the first 5 rows of the dataframe'', etc. If the question is clear enough, please omit this.
\item Use the validator only when necessary. For when and how to use the vaildators, please refer to the examples.
\item Some problemset references are provided below. They are real-world problemsets that are used in data science courses. However, they are not the best examples of problemsets. You are encouraged to write better problemsets than them.
\end{itemize}
\end{tcolorbox}

\section{Problem Examples}
\label{sec:problem-examples}

We provide a few examples for each benchmark here.
The full benchmarks will also become publicly available.

\subsection{DSEval-Kaggle}

We show the first few problems from ``\emph{disease-symptoms-and-patient-profile-dataset}' in DSEval-Kaggle.

\begin{minted}[breaklines,fontsize=\small]{python}
# %%
import pandas as pd
import numpy as np

# %%
"""
query: |
  Import the dataset from `inputs/Disease_symptom_and_patient_profile_dataset.csv`. Assign it to a variable called `disease`.

validator:
  namespace_check:
    disease:
"""

disease = pd.read_csv('inputs/Disease_symptom_and_patient_profile_dataset.csv')

# %%
"""
query: |
  Check the balance of the dataset. Count the number of positive and negative outcomes. Put them in a Series with "Positive" and "Negative" as the index.
"""

disease['Outcome Variable'].value_counts()

# %%
"""
query: |
  Handle the imbalance in the dataset using oversampling. Randomly duplicate some rows from the minority class to make it have the same number of rows as the majority class (use `resample` in sklearn with `random_state` 123 please). Save the balanced dataset in `disease_balanced`.

validator:
  namespace_check:
    disease_balanced:
      ignore_order: true
"""

from sklearn.utils import resample

# Separate majority and minority classes
df_majority = disease[disease['Outcome Variable']=='Positive']
df_minority = disease[disease['Outcome Variable']=='Negative']

# Upsample minority class
df_minority_upsampled = resample(df_minority, 
                                 replace=True,     # sample with replacement
                                 n_samples=df_majority.shape[0],    # to match majority class
                                 random_state=123) # reproducible results

# Combine majority class with upsampled minority class
disease_balanced = pd.concat([df_majority, df_minority_upsampled])

# %%
"""
query: |
  Convert binary features into indicator (0/1) variables, and other categorical features (except "Disease" column) into numerical features using one-hot encoding. Save the encoded dataset in-place.

validator:
  namespace_check:
    disease_balanced:
"""

for column in ['Fever', 'Cough', 'Fatigue', 'Difficulty Breathing']:
    disease_balanced[column] = disease_balanced[column].map({'Yes': 1, 'No': 0})
disease_balanced['Outcome Variable'] = disease_balanced['Outcome Variable'].map({'Positive': 1, 'Negative': 0})

categorical_columns = [column for column in disease_balanced.columns if disease_balanced[column].dtype == 'object' and column != "Disease"]
disease_balanced = pd.get_dummies(disease_balanced, columns=categorical_columns)

# %%
"""
query: |
  Let's assume the name of disease irrelevant for the following case study.
  Split the dataset into training and test sets. The test size should be 20% of the whole dataset. Random state should be set to 42. Use `X_train`, `y_train` to store the training set and `X_test`, `y_test` for test set.

validator:
  namespace_check:
    X_train:
    y_train:
    X_test:
    y_test:
"""

from sklearn.model_selection import train_test_split

X = disease_balanced.drop(['Outcome Variable', 'Disease'], axis=1)
y = disease_balanced['Outcome Variable']

X_train, X_test, y_train, y_test = train_test_split(X, y, test_size=0.2, random_state=42)

# ... more problems omitted
\end{minted}

\subsection{DSEval-Exercise}

Part of problem set ``\emph{02-filtering-sorting-euro12}'' from DSEval-Exercise.

\begin{minted}[breaklines,fontsize=\small]{python}
# %%
import pandas as pd

# %%
"""
query: |
  Import the dataset from `inputs/euro12.csv`.
  Assign it to a variable called euro12.

validator:
  namespace_check:
    euro12:

data:
  euro12.csv: https://raw.githubusercontent.com/guipsamora/pandas_exercises/master/02_Filtering_%26_Sorting/Euro12/Euro_2012_stats_TEAM.csv
"""

euro12 = pd.read_csv('inputs/euro12.csv', sep=',')

# %%
"""
query: Select only the Goal column.
"""

euro12.Goals

# %%
"""
query: How many team participated in the Euro2012?
"""

euro12.shape[0]

# %%
"""
query: What is the number of columns in the dataset?
"""

euro12.info()

# %%
"""
query: View only the columns Team, Yellow Cards and Red Cards and assign them to a dataframe called discipline

validator:
  namespace_check:
    discipline:
"""

discipline = euro12[['Team', 'Yellow Cards', 'Red Cards']]

# ... more problems omitted
\end{minted}

\subsection{DSEval-LeetCode}

Problem ``\emph{duplicate-emails}'' from DSEval-LeetCode.

\begin{minted}[breaklines,fontsize=\small]{python}
# %%

import pandas as pd

# %%

"""
query: |
  Write a function `def duplicate_emails(person: pd.DataFrame) -> pd.DataFrame`.

  `person` is a DataFrame with the following columns:
  - id: int
  - email: str
  `person` contains an email for each record. The emails will not contain uppercase letters.

  The function should return all the duplicate emails. Note that it's guaranteed that the email field is not NULL. Return the result table in **any order**.

  The result format is in the following example.

  Example input:
  ```
  person:
  +----+---------+
  | id | email   |
  +----+---------+
  | 1  | a@b.com |
  | 2  | c@d.com |
  | 3  | a@b.com |
  +----+---------+
  ```

  Example output:
  ```
  +---------+
  | email   |
  +---------+
  | a@b.com |
  +---------+
  ```

  Example explanation: a@b.com is repeated two times.

validator:
  table_test:
    function_name: duplicate_emails
    input_validator: |
      def _validate(person):
        assert person.shape[0] > 0
        assert person.dtypes.equals(pd.Series({'id': 'int64', 'email': 'object'}))
        assert person.id.is_unique
        assert person.email.str.match(r'^[a-z0-9._%+-]+@[a-z0-9.-]+\.[a-z]{2,}$').all()

    output_checker:
      ignore_order: true

    test_cases:
    - # example
      - "`pd.DataFrame({'id': [1, 2, 3], 'email': ['a@b.com', 'c@d.com', 'a@b.com']})`"
    - # corner case: only one email
      - "`pd.DataFrame({'id': [1], 'email': ['a@b.com']})`"
    - # corner case: all emails are the same
      - "`pd.DataFrame({'id': [1, 2, 3], 'email': ['a@b.com', 'a@b.com', 'a@b.com']})`"
    - # corner case: all emails are different
      - "`pd.DataFrame({'id': [1, 2, 3], 'email': ['a@b.com', 'c@d.com', 'e@f.com']})`"
    - # corner case: some emails are the same
      - "`pd.DataFrame({'id': [1, 2, 3, 4], 'email': ['a@b.com', 'c@d.com', 'a@b.com', 'c@d.com']})`"
    - # corner case: some emails are the same, but not all
      - "`pd.DataFrame({'id': [1, 2, 3, 4, 5], 'email': ['a@b.com', 'c@d.com', 'a@b.com', 'c@d.com', 'e@f.com']})`"
"""

def duplicate_emails(person: pd.DataFrame) -> pd.DataFrame:
    # Group by email and count the occurrences
    email_counts = person.groupby("email").size().reset_index(name="count")

    # Filter the emails with count greater than 1 (duplicates)
    duplicates = email_counts[email_counts["count"] > 1]

    # Return the duplicate emails as a DataFrame
    return duplicates[["email"]]
\end{minted}

\subsection{DSEval-SO}

Problem ``\emph{numpyeval-001}'' from DSEval-SO.

\begin{minted}[breaklines,fontsize=\small]{python}
# %%
import numpy as np
from numpy import newaxis

a = np.array([[1, 2, 3], [3, 4, 5], [5, 6, 7]])

# %%
"""
query: |
  I have a 2d array with shape (x, y) which I want to convert to a 3d array with shape (x, y, 1).
  Is there a nice Pythonic way to do this?
"""

a[:, :, newaxis]
\end{minted}

\end{document}